\documentclass{article}
\PassOptionsToPackage{numbers, compress}{natbib}

\usepackage[final]{neurips_2022}

\usepackage[utf8]{inputenc} % allow utf-8 input
\usepackage[T1]{fontenc}    % use 8-bit T1 fonts
\usepackage[pagebackref=true,breaklinks=true,letterpaper=true,colorlinks,bookmarks=false]{hyperref}       % hyperlinks
\usepackage{url}            % simple URL typesetting
\usepackage{booktabs}       % professional-quality tables
\usepackage{amsfonts}       % blackboard math symbols
\usepackage{nicefrac}       % compact symbols for 1/2, etc.
\usepackage{microtype}      % microtypography
\usepackage{xcolor}         % colors
\usepackage{xspace}
\usepackage{amsmath}

\usepackage{url}            % simple URL typesetting
\usepackage{cellspace}       % professional-quality tables
\usepackage{nicefrac}       % compact symbols for 1/2, etc.
\usepackage{microtype}      % microtypography
\usepackage{adjustbox}
\usepackage{parskip}
\usepackage{float}
\usepackage{mathtools}
\usepackage{wrapfig}
\usepackage[normalem]{ulem}
\usepackage{listings}

\usepackage{algorithm}
\usepackage{algpseudocode}
\algrenewcommand\algorithmicrequire{\textbf{Input:}}
\algrenewcommand\algorithmicensure{\textbf{Output:}}

\DeclareMathOperator*{\argmin}{arg\,min}

\newcommand{\lorl}{LOReL\xspace}
\newcommand{\LISA}{LISA\xspace}
\newcommand{\X}{LISA\xspace}

\definecolor{codegreen}{rgb}{0,0.6,0}
\definecolor{codegray}{rgb}{0.5,0.5,0.5}
\definecolor{codepurple}{rgb}{0.58,0,0.82}
\definecolor{backcolour}{rgb}{0.95,0.95,0.92}

\lstdefinestyle{mystyle}{
    backgroundcolor=\color{backcolour},   
    commentstyle=\color{codegreen},
    keywordstyle=\color{magenta},
    numberstyle=\tiny\color{codegray},
    stringstyle=\color{codepurple},
    basicstyle=\ttfamily\footnotesize,
    breakatwhitespace=false,         
    breaklines=true,                 
    captionpos=b,                    
    keepspaces=true,                 
    numbers=left,                    
    numbersep=5pt,                  
    showspaces=false,                
    showstringspaces=false,
    showtabs=false,                  
    tabsize=2
}

\makeatletter
\newcommand{\printfnsymbol}[1]{%
  \textsuperscript{\@fnsymbol{#1}}%
}
\makeatother

\title{LISA: Learning Interpretable Skill Abstractions from Language}

\author{%
  Divyansh Garg$^*$ \hspace{10pt} 
  Skanda Vaidyanath$^*$ \hspace{10pt}
  Kuno Kim \\
  \textbf{Jiaming Song} \hspace{10pt}
  \textbf{Stefano Ermon} \\
  Stanford University \\
  \texttt{\{divgarg, svaidyan, khkim, tsong, ermon\} @stanford.edu} \\
}

\begin{document}

\def\thefootnote{*}\footnotetext{Equal contribution.}
\renewcommand*{\thefootnote}{\arabic{footnote}}
\setcounter{footnote}{0}

\maketitle

\begin{abstract}

Learning policies that effectively utilize language instructions in complex, multi-task environments is an important problem in sequential decision-making. While it is possible to condition on the entire language instruction directly, such an approach could suffer from generalization issues. In our work, we propose \emph{Learning Interpretable Skill Abstractions (LISA)}, a hierarchical imitation learning framework that can learn diverse, interpretable \emph{primitive behaviors} 
 or skills from language-conditioned demonstrations to better generalize to unseen instructions. LISA uses vector quantization to learn discrete skill codes that are highly correlated with language instructions and the behavior of the learned policy. In navigation and robotic manipulation tasks, LISA outperforms a strong non-hierarchical Decision Transformer baseline in the low data regime and is able to compose learned skills to solve tasks containing unseen long-range instructions. Our method demonstrates a more natural way to condition on language in sequential decision-making problems and achieve interpretable and controllable behavior with the learned skills.
\end{abstract}
\section{Introduction}
\label{intro}

\begin{wrapfigure}{r}{0.45\textwidth}
  %\vskip -10pt
  %\centering
  %\fbox{\rule[-.5cm]{0cm}{4cm} \rule[-.5cm]{4cm}{0cm}}
  \hspace*{-0.1cm}\includegraphics[width=\linewidth]{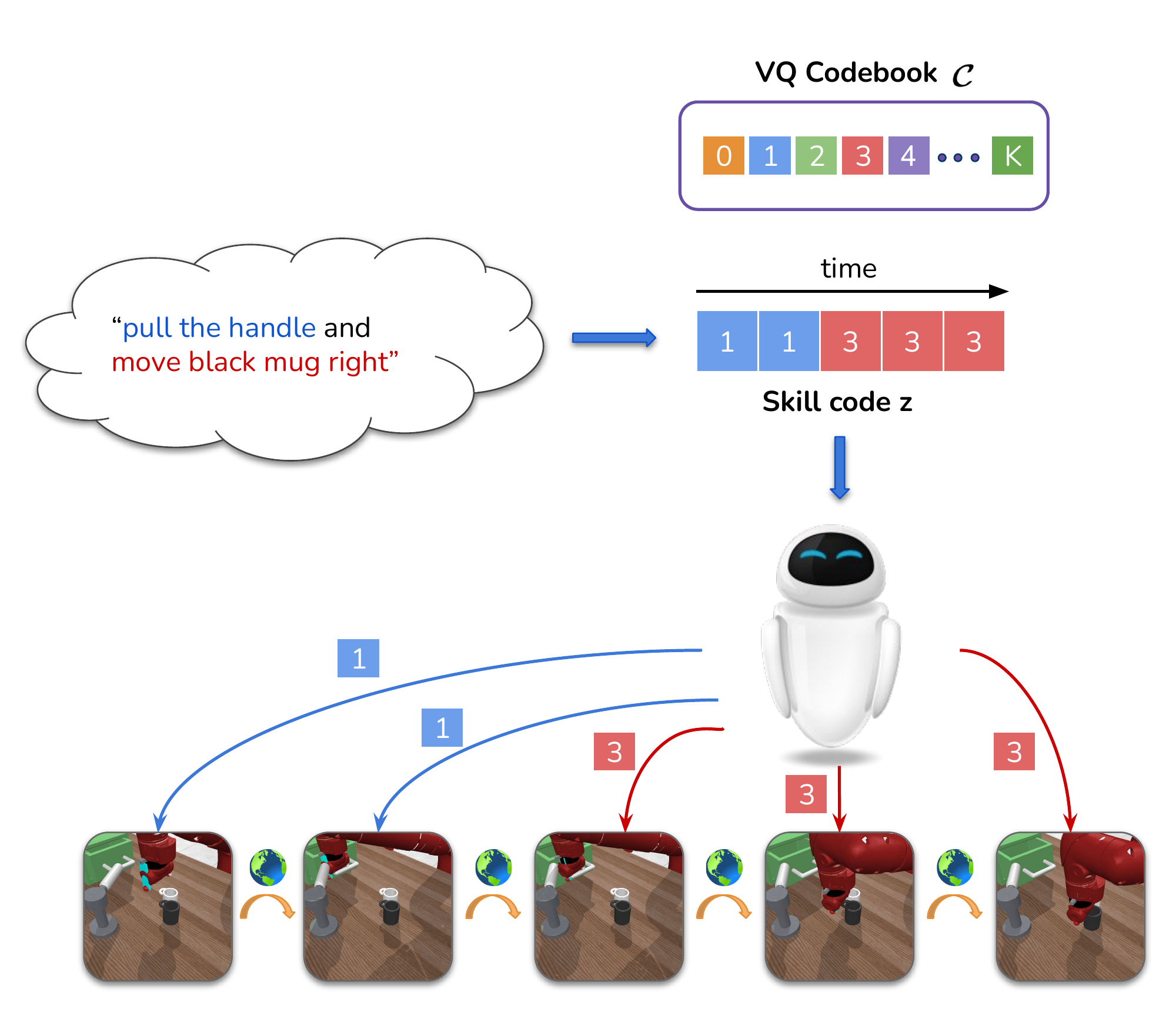}
  \vskip -5pt
  \caption[]
  {\small\textbf{Overview of \LISA.} Given a language instruction, our method learns discrete skill abstractions $z$, picked from a vector codebook $\mathcal{C}$. The policy conditioned on a skill executes distinct behaviors and solves different sub-tasks. See \href{https://gifyu.com/image/Sbtpm}{GIF}.}
  \label{fig:intro}
  \vskip -10pt
\end{wrapfigure}

Intelligent machines should be able to solve a variety of complex, long-horizon tasks in an environment and generalize to novel scenarios. 
In the sequential decision-making paradigm, provided expert demonstrations, an agent can learn to perform these tasks via multi-task imitation learning (IL).
As humans, it is desirable to specify tasks to an agent using a convenient, yet expressive modality and the agent should solve the task by taking actions in the environment.
There are several ways for humans to specify tasks to an agent, such as task IDs, goal images, and goal demonstrations. However, these specifications tend to be ambiguous, require significant human effort, and can be cumbersome to curate and provide at test time. 
One of the most natural and versatile ways for humans to specify tasks is via natural language.

The goal of language-conditioned IL is to solve tasks in an environment given language-conditioned trajectories at training time and a natural language instruction at test time.
This becomes challenging when the task involves completing several sub-tasks sequentially, like the example shown in Figure \ref{fig:intro}.
A crucial step towards solving this problem is exploiting the inherent hierarchical structure of natural language.
For example, given the task specification ``pull the handle and move black mug right'', we can split it into learning two independent \textit{primitive behaviors or skills}, \textit{i.e.} ``pull the handle'' and ``move black mug right''.
If we are able to decompose the problem of solving these complex tasks into learning skills, we can re-use and compose these learned skills to generalize to unseen tasks in the future.
This is especially useful in the low-data regime, since \textit{we may not see all possible tasks given the limited dataset, but may see all the constituent sub-tasks.}
Using such hierarchical learning, we can utilize language effectively and learn skills as the building blocks of complex behaviors. 

Utilizing language effectively to learn skills is a non-trivial problem and raises several challenges.
(i) The process of learning skills from language-conditioned trajectories is unsupervised as we may not have knowledge about which parts of the trajectory corresponds to each skill.
(ii) We need to ensure that the learned skills are useful, \textit{i.e.} encode behavior that can be composed to solve new tasks.
(iii) We would like the learned skills to be interpretable by humans, both in terms of the language and the behaviours they encode. 
There are several benefits of interpretability. For example, it allows us to understand which skills our model is good at and which skills it struggles with. In safety critical settings such as robotic surgery or autonomous driving, knowing what each skill does allows us to pick and choose which skills we want to run at test time. It also provides a visual window into a neural network policy which is extremely desirable \cite{vizbert}.
There have been prior works such as \cite{lorl, cliport, babyai} that have failed to address these challenges and condition on language in a monolithic fashion without learning skills. As a result, they tend to perform poorly on long-horizon composition tasks such as the one in Figure \ref{fig:intro}.

\begin{figure}[t]
  \centering
  %\fbox{\rule[-.5cm]{0cm}{4cm} \rule[-.5cm]{4cm}{0cm}}
  \hspace*{-2cm}\includegraphics[width=56 em]{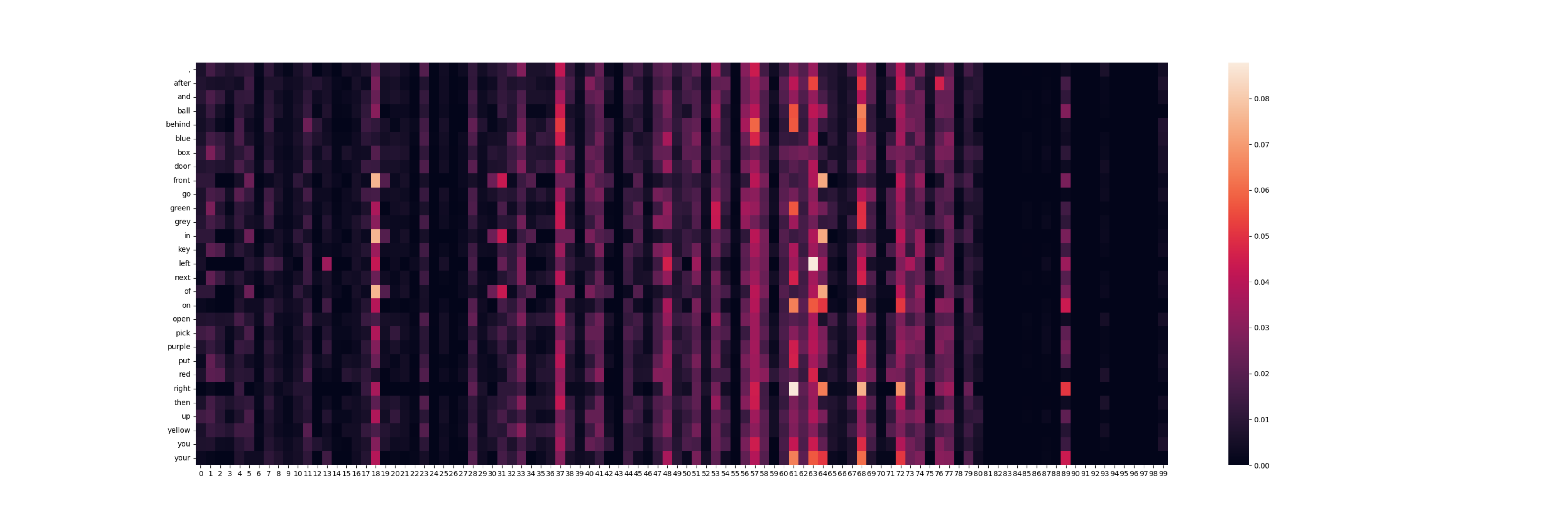}
  \vskip -20pt
  \caption{\small \textbf{\X Skill Heat map.} We show the corresponding word frequencies for 100 different learned skills on BabyAI BossLevel task (column normalized). The x-axis is the skill index and the y-axis is the task  vocabulary. LISA's learned skills are interpretable, encode diverse behavior, and are distinctly activated for different words. (zoom in for the best view) }
  \label{fig:boss_example}
  \vskip -5pt
\end{figure}

To this end, we propose \textbf{L}earning \textbf{I}nterpretable \textbf{S}kill \textbf{A}bstractions from language \textbf{(LISA), a hierarchical imitation learning framework that can learn interpretable skills from language-conditioned offline demonstrations.}
\X uses a two-level architecture -- a skill predictor that predicts quantized skills from a learnt vector codebook and a policy that uses these skill vector codes to predict actions. The discrete skills learned from language are interpretable (see Figure \ref{fig:boss_example} and \ref{fig:word_cloud_gif}) and can be composed to solve long-range tasks. 
Using quantization maximizes skill reuse and enforces a bottleneck to pass information from the language to the policy, enabling unsupervised learning of interpretable skills. 
We perform experiments on grid world navigation and robotic manipulation tasks and show that our hierarchical method can outperform a strong non-hierarchical baseline based on Decision Transformer~\cite{decision-transformer} in the low-data regime.
We analyse these skills qualitatively and quantitatively and find them to be highly correlated to language and behaviour. 
Finally, using these skills to perform long-range composition tasks on a robotic manipulation environment results in performance that is nearly \emph{2x better} than the non-hierarchical version.

Concretely, our contributions are as follows:

\begin{itemize}
\item We introduce \X, a novel hierarchical imitation framework to solve complex tasks specified via language by learning re-usable skills. 
\item We demonstrate the effectiveness of our approach in the low-data regime where its crucial to break down complex tasks to generalize well.
\item We show our method performs well in long-range composition tasks where we may need to apply multiple skills sequentially.
\item We also show that the learned skills are highly correlated to language and behaviour and can easily be interpreted by humans.
\end{itemize}

\section{Related Work}
\label{related}

\subsection{Imitation Learning}
Imitation learning (IL) has a long history, with early works using behavioral cloning~\cite{pomerleau1991efficient, ross2010efficient, ross2011reduction}
to learn policies via supervised learning on expert demonstration data.
Recent methods have shown significant improvements via learning reward functions~\cite{gail} or Q-functions~\cite{iq} from expert data to mimic expert behavior. Nevertheless, these works typically consider a single task. 
An important problem here is multi-task IL, where the imitator is trained to mimic behavior on a variety of training tasks with the goal of generalizing the learned behaviors to test tasks. A crucial variable in the multi-task IL set-up is \emph{how the task is specified}, e.g vectorized representations of goal states \cite{nair2018visual}, task IDs \cite{kalashnikov2021mtopt}, and single demonstrations \cite{xu2018neural, duan2017one, finn2017oneshot, yu2018oneshot}. In contrast, we focus on a multi-task IL setup with task-specification through language, one of the most natural and versatile ways for humans to communicate desired goals and intents.

\subsection{Language Grounding}
Several prior works have attempted to ground language with tasks or use language as a source of instructions for learning tasks with varying degrees of success~\cite{10.5555/1597348.1597423, wang2016learning, 2017, oh2017zeroshot, 10.1007/s10514-018-9792-8}. \cite{luketina2019survey} is a good reference for works combining language with sequential-decision making.

But apart from a few exceptions, most algorithms in this area use the language instruction in a monolithic fashion and are designed to work for simple goals that requires the agent to demonstrate a single skill~\cite{paxton2019prospection, chaplot2018gatedattention, hermann2017grounded, blukis2018following} or tasks where each constituent sub-goal has to be explicitly specified~\cite{chen2020touchdown, alfred, misra2017mapping, anderson2018visionandlanguage, tan2019learning, suhr2018situated, fried2018speakerfollower, misra2019mapping, ma2019selfmonitoring}. 
Some recent works have shown success on using play data~\cite{lang-play} or \emph{pseudo-expert} data such as \lorl~\cite{lorl} and CLIPORT~\cite{cliport}. \lorl and CLIPORT are not hierarchical techniques. \cite{lang-play} can be interpreted as a hierarchical technique that generates latent sub-goals as a function of goal images, language instructions and task IDs but the skills learned by \X are purely a function of language and states alone and do not require goal images or task IDs.
\cite{lah, hu2019hierarchical} and \cite{stepputtis2020languageconditioned} are some examples of works that use a two-level architecture for language conditioned tasks but neither of these methods learn skills that are interpretable.
%\cite{lah} is a hierarchical RL method that uses a high-level policy conditioned on the state that generates language instructions for the low-level policy to follow. The low-level policy has direct access to the language instruction while acting in the environment. 
%  \se{this is hard to understand without knowing what the method is/does..shuld related work section be moved later?}
% \dg{could make sense}

\subsection{Latent-models and Hierarchical Learning}
Past works have attempted to learn policies conditioned on latent variables and some of them can be interpreted as hierarchical techniques. 
For example, \cite{diayn} learns skills using latent variables that visit different parts of the environment's state space. \cite{dads} improved on this by learning skills that were more easily predictable using a dynamics model. But these fall more under the category of skill discovery than hierarchical techniques since the skill code is fixed for the entire trajectory, as is the case with \cite{diayn}. 
\cite{latent-play} and \cite{infogail} are other works that use a latent-variable approach to IL. But these approaches don't necessarily learn a latent variable with the intention of breaking down complex tasks into skills.
With \X, we sample several skills per trajectory with the clear intention of each skill corresponding to completing a sub-task for the whole trajectory.
Also, none of the methods mentioned here condition on language.
% \se{need to state a few difference before going into RL and options?} 

There has been some work on hierarchical frameworks for RL to learn high-level action abstractions, called \textit{options}~\cite{options}, such as
\cite{li2020subpolicy, zhang2021hierarchical, nachum2018dataefficient} but these works are not goal-conditioned. Unlike \X, these works don't use language and the options might lack diversity and not correspond to any concrete or interpretable skills. Furthermore, none have used the VQ technique to learn options and often suffer from training instabilities. 

\section{Approach}
\label{method}

% \js{what is the issue that you are trying to address again? is there a short sentence that can distill the key idea? such as ``The key idea of LISA is to learn quantized skill representations that are informative to both language and behaviors, which allows us to do this do that and that.}

The key idea of LISA is to learn quantized skill representations that are informative of both language and behaviors,
which allows us to break down high-level instructions, specified via language, into discrete, interpretable and composable codes (see Fig.~\ref{fig:boss_example}, Fig.~\ref{fig:word_clouds} and Fig.~\ref{fig:boss_heat} for visualizations). These codes enable learning explainable and controllable behaviour, as shown in Fig.~\ref{fig:intro} and Fig.~\ref{fig:word_cloud_gif}.

% \se{following two paragraphs can be cut i think, won't make much sense bfore reading the rest} 
% \X consists of two modules trained end-to-end: (1) a high-level skill predictor, acting as a meta-controller, and (2) a low-level policy. 

% The first stage learns the quantized skill codes from language,  and the second stage uses the codes to obtain diverse behavior.

% We consider an unsupervised RL paradigm in this work, where the agent is allowed an unsupervised
% “exploration” stage followed by a supervised stage.

% Conveniently, because skills are learned without a priori knowledge of the task, the learned skills can be used for many different tasks.

Section \ref{sec:method} describes the problem formulation, an overview of our framework, and presents our language-conditioned model. Section \ref{sec:training} provides details on the training approach.

%Our approach is summarized in Algorithm~\ref{alg:lisa}

% language, traj --> option selector --> discrete option code --> states, actions, option --> decision transformer --> imitate for H steps\\
% no language going into DT, acts as bottle neck -- forces options to be lang dependant\\
% discrete option code comes from VQ-VAE codebook \\

\DeclarePairedDelimiter\floor{\lfloor}{\rfloor}

\subsection{Language-conditioned Skill Learning}
\label{sec:method}

\subsubsection{Problem Setup}

We consider general multi-task environments, represented as a task-augmented Markov decision process (MDP) with a family of different tasks $\mathcal{T}$. A task $\mathcal{T}_i$ may be composed of other tasks in $\mathcal{T}$ and encode multiple sub-goals.
For example, in a navigation environment, a task could be composed of two or more sub-tasks - ``pick up ball'', ``open door'' -  in any hierarchical order.
$\mathcal{S}, \mathcal{A}$ represent state and action spaces. We assume that each full task has a \emph{single} natural language description $l \in L$, where $L$ represents the space of language instructions.
Any sub-goals for the task are encoded within this single language instruction.
% Moreover, the language description $l$ is presented to an agent at the start of an episode,  
% Any $l$ can be composed of other language instructions using a specific grammer (say English) in $L$.

% We define a probabilistic context variable denoted as $z \in \mathcal{Z}$, where $\mathcal{Z}$ is the (discrete or continuous) value space of $Z$.
%  For example, in a navigation task, the context variables could represent different goal positions in the environment. 
% Now, each component of the MDP has an additional dependency on the context variable $m$.

We assume access to an offline dataset $\mathcal{D}$ of trajectories obtained from an optimal policy for a variety of tasks in an environment with only their language description available.
% We consider an imitation learning setup with access to a dataset $\mathcal{D} = \{(l_1, \tau_1), (l_2, \tau_2), ..., (l_N, \tau_N)\}$ consisting of $N$ expert trajectories $\tau_i$ paired with a language instruction $l_i \in L$.
Each trajectory $\tau^i = (l^i, \{(s^i_1, a^i_1), (s^i_2, a^i_2), ..., (s^i_T, a^i_T)\}$) consists of the language description and the observations $s^i_t \in \mathcal{S}$, actions $a^i_t \in \mathcal{A}$ taken over $T$ timesteps. The trajectories are not labeled with any rewards.

Our aim is to predict the expert actions  $a_t$, given a language instruction and past observations.
% \sout{Unlike some prior works, we assume a \emph{single language instruction} for the whole task, \textit{e.g.}, \textit{"Go to the red box and pick up the green ball"}, instead of separate instructions for each sub-goal.} 

Note that each trajectory in the training dataset can comprise of any number of sub-tasks. 
For example, we could have a trajectory to ``open a door'' and another to ``pick up a ball and close the door'' in the training data. 
With \X we aim to solve the task ``open a door and pick up the ball'' at test time even though we haven't seen this task at training time.
In a trajectory with multiple sub-tasks, the training dataset \textbf{does not} give us information about where one sub-task ends and where another one begins.

% If we had this information, we could simply learn the trajectory corresponding to each task independently and run them sequentially at test time depending on the instruction.
% Therefore, we can think of \X as performing unsupervised sub-task identification with natural language, which is a challenging problem on its own.
\X must learn how to identify and stitch together these sub-tasks learned during training, in order to solve a new language instruction such as the one shown in Fig.~\ref{fig:intro} at test time.

\begin{wrapfigure}{r}{0.5\textwidth}
  \vskip -10pt
  %\fbox{\rule[-.5cm]{0cm}{4cm} \rule[-.5cm]{4cm}{0cm}}
  \hspace*{-0.1cm}\includegraphics[width=\linewidth]{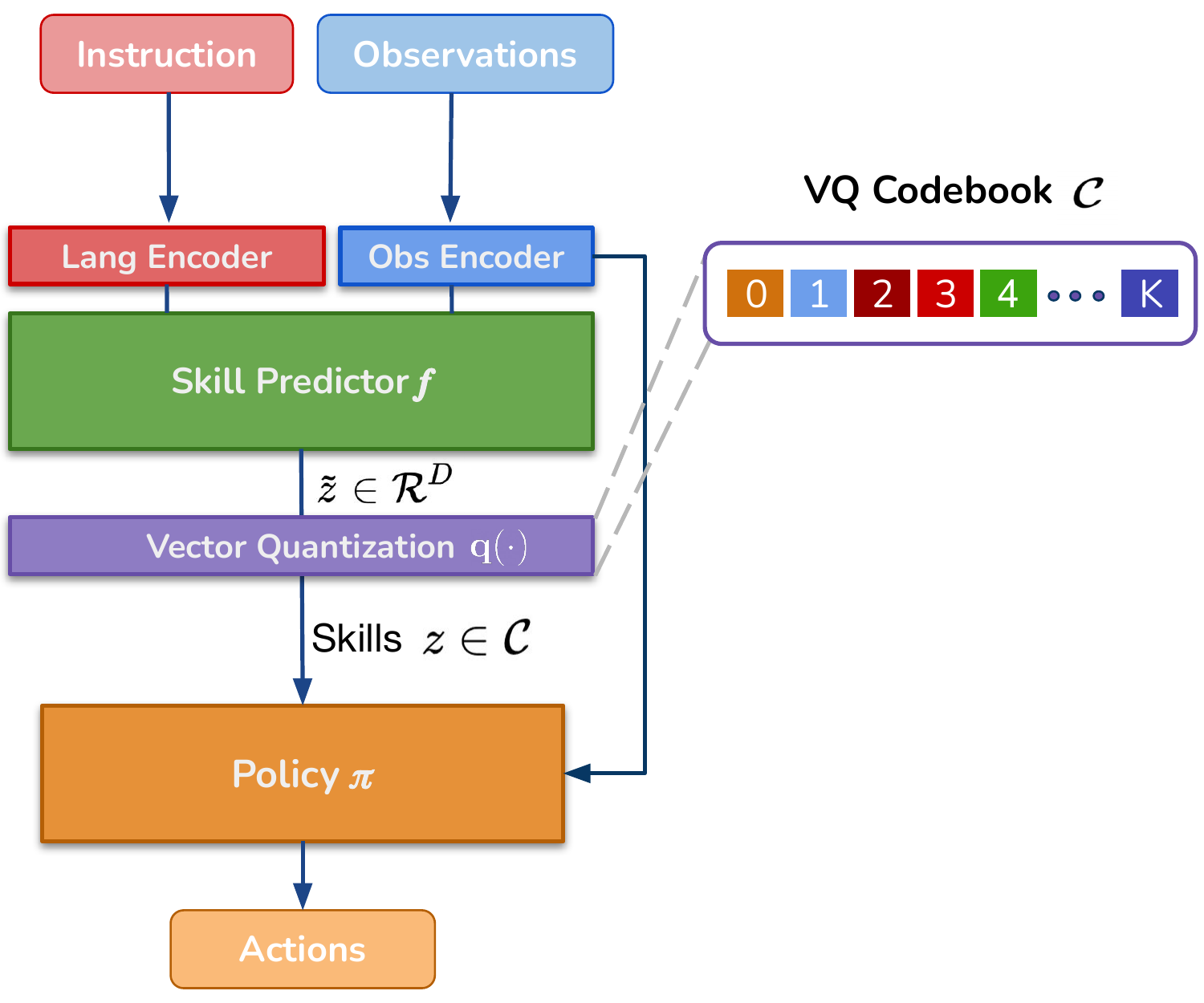}
  \vskip -5pt
 \caption{\small \textbf{\X Architecture}: The skill predictor $f$ gets the language instruction and a sequence of observations as the input, processed through individual encoders. It predicts quantized skill codes $z$ using a learnable cookbook $\mathcal{C}$, that encodes different sub-goals, and passes them to the policy $\pi$. \X is trained end-to-end.}
  \label{fig:method}
  \vskip -20pt
\end{wrapfigure}

\subsubsection{Hierarchical Skill Abstractions}
We visualize the working of \X in Figure~\ref{fig:method}. Our framework consists of two modules: a skill predictor $f: L \times  \mathcal{S} \rightarrow \mathcal{C}$ and a policy $\pi: \mathcal{S} \times \mathcal{C} \rightarrow \mathcal{A}$. 
% \se{i thought you wanted history dependent policy} 
Here, $\mathcal{C} = \left\{z^{1}, \ldots, z^{K}\right\}$ is a learnable codebook of $K$ quantized skill latent codes. $D$ is the dimension of the latent space of skills.

% \js{domain and co-domains}

% \js{is the skill predictor trained end-to-end?}\dg{Yes, its trained in conjunction with the policy, we are not pre-training it}

Our key idea is to break learning behavior from language in two stages: 1) Learn  discrete latent codes $z$, representing skills, from the full-language instruction to decompose the task into smaller sub-goals 2) Learn a policy $\pi$ conditioned only on these discrete codes. In \X, both stages are trained end-to-end.

% \js{this makes it sound like it is no longer end-to-end}

Given an input $\tau = (l, \{s_t, a_t\}_{t=1}^T)$, the skill predictor $f$ predicts a skill code $\tilde{z} \in \mathcal{R}^D$ at a timestep $t$  as $\tilde{z} = f(l, (s_t, s_{t-1}, ...))$.  These codes are discretized using a vector quantization operation $\mathbf{q(\cdot)}$ that maps a latent $\tilde{z}$ to its closest codebook entry $z = \mathbf{q}({\tilde{z}})$.  The quantization operation $\mathbf{q(\cdot)}$  helps in learning discrete skill codes and acts as a bottleneck on passing language information. We detail its operation in  Sec.~\ref{sec:training}.

The chosen skill code $z$, is persisted for $H$ timesteps where $H$ is called the horizon. More details on how we chose the horizon and ablations studies on the choice of $H$ can be found in appendix sections \ref{sec:training_deets} and \ref{sec:ablations}. After $H$ timesteps, the skill predictor is invoked again to predict a new skill. This enforces the skill to act as a temporal abstraction on actions, i.e. options~\cite{options}.
% Given an input $\tau = (l, \{s_t, a_t\}_{t=1}^T)$, the skill predictor $f$ encodes $l$ as skill codes, conditioned on the seen states  $H$ timesteps, where $H$ is the horizon for which we persist the skill. $\tilde{z_k} = f(l \mid (s_k, s_{k-1}, ...))$ for $k \in {H}$. These codes are discretized using a vector quantization operation $\mathbf{q(\cdot)}$ that maps a code $\tilde{z}$ to its closest codebook entry $z = \mathbf{q}({\tilde{z}})$.
% to be one of $K$ vectors in a codebook $\mathcal{C}$ as $z_t^\mathrm{q} = \mathrm{\mathbf{q}}(z_t)$.
% \js{VQ is not defined.} 
The policy $\pi$  predicts the action $a_t$ at each timestep $t$ conditioned on the state and a single skill code  $z$ that is active at that timestep. For $\pi$ to correctly predict the original actions, it needs to use the language information encoded in the skill codes.
% $\pi$ samples new skills codes only every horizon $H$ steps, reusing them for this duration. These make the skill act as a temporal abstraction, i.e. options.

% \js{technically, you are not actually reconstructing observations here, so VQ objective does not apply directly, why not directly write the objective that you are actually using?}
% The learnt skills are discretized using Vector Quantization (VQ) and act as a bottleneck on passing language information to the downstream policy $\pi$. \js{need to discuss VQ in detail, either here or in background.} For $\pi$ to correctly imitate an expert, it needs to be able to use the language-information in the skills to predict the correct actions. 

% \sv{why discrete codes and not continuous}
\LISA learns quantized skill codes in a vector codebook instead of continuous embeddings as this encourages reusing and composing these codes together to pass information from the language input to the actual behavior. Our learnt discrete skill codes adds interpretability and controllability to the policy's behavior.

\subsection{Training LISA}
\label{sec:training}

\paragraph{Learning Discrete Skills.} 
% \subsubsection{Compression with Vector Quantization}
% \se{this subsection feels out of place, breaks the flow}
% \se{add citations to the older stuff on dictionary learning}
 \X uses Vector Quantization (VQ), inspired from \cite{vq-vae}. It is a natural and widely-used method to map an input signal to a low-dimensional discrete learnt representation. VQ learns a codebook $\mathcal{C} \in\left\{z^{1}, \ldots, z^{K}\right\}$ of $K$ embedding vectors. Given an embedding $\tilde{z}$ from the skill predictor $f$, it maps the embedding to the closest vector in the codebook:
$$
z = \mathbf{q}(\tilde{z}) =: \argmin _{z^k \in \mathcal{C}}\|\tilde{z}- z^k\|_{2}
$$
with the codebook vectors updated to be the moving average of the embeddings $z$ closest to them.
 This can be classically seen as learning $K$ cluster centers via $k$-means~\cite{vq}.  
% The signal $X$ can be reconstructed using the decoder as $\Tilde{X} =  D(\vct{z_q})$ to approximate $X$.

Backpropagation through the non-differentiable quantization operation is achieved by a straight-through gradient estimator, which simply copies the gradients from the decoder to the encoder, such that the model and codebook can be trained end-to-end.

% The skill predictor $f$ learns skills $\vct{z_q}^t$ taking the language $l$ and sequence of observations $o$ as input. 
VQ enforces each learnt skill $z$ to lie in $\mathcal{C}$,  which can be thought as learning $K$ prototypes or cluster centers for the language embeddings using the seen states. This acts as a bottleneck that efficiently decomposes a language instruction into sub-parts encoded as discrete skills.

\textbf{\X Objective.} \X is trained end-to-end using an objective $\mathcal{L}_\mathrm{LISA} = \mathcal{L}_\mathrm{BC} + \lambda \mathcal{L}_\mathrm{VQ}$, where $\mathcal{L}_\mathrm{BC}$ is the behavior-cloning loss on the policy $\pi_\theta$, $\lambda$ is the VQ loss weight and $\mathcal{L}_\mathrm{VQ}$ is the vector quantization loss on the skill predictor $f_\phi$ given as:
\begin{equation}
\mathcal{L}_{\mathrm{VQ}}(f)=\mathbb{E}_{\tau}[\left\|\operatorname{sg}\left[\mathbf{q}(\tilde{z})\right]-\tilde{z}\right\|_{2}^{2}] 
\end{equation}
with $\tilde{z} = f_\phi(l, (s_t, s_{t-1}, ..))$. 

% \js{missing parenthesis? maybe remind reader where l comes from (tau)}
% \js{should have an expectation over langauge and behaviors? how are f and c used in the above expression?}
% Here,  $\mathcal{L}$ is the cross-entropy loss for discrete $\mathcal{A}$ and MSE loss for continuous $\mathcal{A}$.
sg $[\cdot]$ denotes the stop-gradient operation. $\mathcal{L}_{\mathrm{VQ}}$ is also called \emph{commitment loss}.
It minimizes the conditional entropy of the skill predictor embeddings given the codebook vectors, making the embeddings stick to a single codebook vector.

The codebook vectors are learnt using an exponential moving average update, same as~\cite{vq-vae}.

\begin{algorithm}[t]
   \caption{Training LISA}
   \label{alg:lisa}
\begin{algorithmic}[1]
   \Require Dataset $\mathcal{D}$ of language-paired trajectories
    \Require Num skills $K$ and horizon $H$
   \State Initialize skill predictor $f_\phi$, policy $\pi_\theta$
   \State Vector Quantization op $\mathbf{q(\cdot)}$
  \While {\textit{not converged}}

   \State{Sample $\tau = (l, \{s_0, s_1, s_2...s_T\}, \{a_0, a_1, a_2...a_T\}$)}
   \State Initialize $S=\{s_0\}$ \Comment{List of seen states}
   \For{$k=0..\floor*{\frac{T}{H}}$} \Comment{Sample a skill every H steps}

%   \State $current\_traj = [s_0]$ \Comment{Initialize with start state}
%   \State  Append $s_0$ to $S$ \Comment{Add initial state}
%   \State $action\_preds = []$
%   \State $k=0$
%   \Repeat 
   \State  $z \leftarrow \mathbf{q}(f_\phi(l, S))$
   \For{\textit{step} $t=1..H$}
   \Comment{Predict actions using a fixed skill and context length $H$}
   \State $a_{k H+t} \leftarrow \pi_\theta(z, S[:-H])$ % \Comment{Use c $H$ steps}
    \State  $S \leftarrow S \cup \{s_{k H+t}\} $ \Comment{Append seen state}
%   \State $action\_preds.append(a)$
%   \State $current\_traj.append(A[k*H+i])$
%   \State $current\_traj.append(S[k*H+i+1])$
%   \State $states.append(S[k*H+i+1])$
   \EndFor
%   \State $k+=1$
%   \Until{end of trajectory}
   \State Train $f_\phi, \pi_\theta$ using objective $\mathcal{L_\mathrm{LISA}}$
   \EndFor
   \EndWhile
\end{algorithmic}
\end{algorithm}

\textbf{Avoiding language reconstruction.}  \LISA avoids auxiliary losses for language reconstruction from the skills latent codes and it's not obvious why the skill codes are properly encoding language, and we expand on it here.
%or self-supervised learning. 
%As noted, \LISA works similar to MAE, nevertheless, , when trained without any language supervision.
% where part of an input (here the actions and future states) is hidden and can be reconstructed to learn a latent representation.

For a given a signal $X$ and a code $Z$, reconstructing the signal from the code as $\tilde{X}=f(Z)$ using cross-entropy loss amounts to maximizing the Mutual Information (MI) $I(X, Z)$ between $X$ and $Z$ \cite{vim, MI}.  In our case, we can write the MI between the skill codes and language using entropies as: $I(z, l) = H(z) - H(z \mid l)$, whereas methods that attempt to reconstruct language apply the following decomposition: $I(z, l) = H(l) - H(l \mid z)$. Here, $H(l)$, the entropy of language instructions, is constant, and this gives us the cross-entropy loss.

% Now maximizing $I(z, l)$ is the same as minimizing the cross-entropy between the actual language and language reconstruction using the skills.
Thus we can avoid language reconstruction via cross-entropy loss by maximizing $I(z, l)$ directly. In \LISA, $\mathcal{L}_\mathrm{vq} = -H(z \mid l)$, and we find there is no need to place a constraint on $H(z)$ as the learned skill codes are diverse, needing to encode enough information to correctly predict the correct actions.\footnote[1]{In experiments, we tried enforcing a constraint on $H(z)$ by using extra InfoNCE loss term but don't observe any gains.}

% \sv{Lets not talk about every random thing we tried, its getting very confusing. In the last three paragraphs, we have self-attention, MAE, InfoNCE, VQ, reconstruction -- wayyyyy too much going on}

As a result, \LISA can maximize the MI between the learnt skills and languages without auxiliary reconstruction losses and enforcing only $\mathcal{L}_\mathrm{vq}$ on the skill codes. We empirically estimate the MI between the language and skill codes and find that our experiments confirm this in Section~\ref{sec:MI}.

\begin{figure*}[t]
  \vskip -15 pt
  \centering
  %\fbox{\rule[-.5cm]{0cm}{4cm} \rule[-.5cm]{4cm}{0cm}}
  \includegraphics[width=\linewidth]{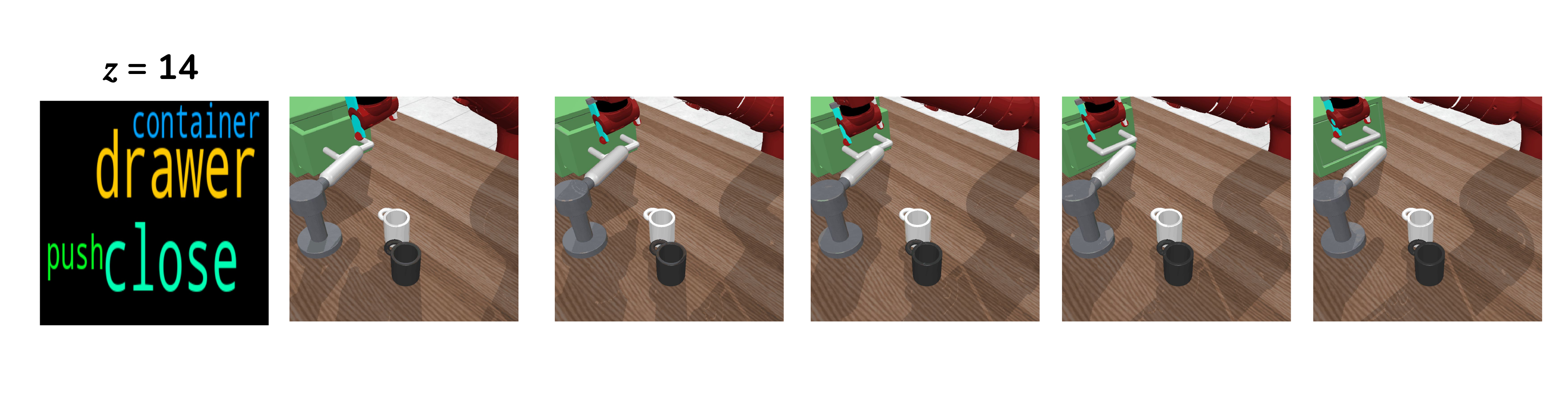}
  \vskip -30 pt
  \caption{\small\textbf{Behavior with fixed LISA options.} We show the word clouds and the behavior of the policy obtained by using a fixed skill code $z = 14$ for an entire episode. We find that this code encodes the skill ``closing the drawer'', as indicated by the word cloud. The policy executes this skill with a high degree of success when conditioned on this code for the entire trajectory, across multiple environment initializations and seeds.}
  \label{fig:word_cloud_gif}
  \vspace{-10pt}
\end{figure*}

\subsubsection{LISA Implementation}
\label{sec:flat}

% \sv{Talk about how there are two transformers but still compute isnt that much more than vanilla equivalent}
\X can be be implemented using different network architectures, such as Transformers or MLPs. 

In our experiments, we use Transformer architectures with \LISA, but we find that out method is effective even with simple architectures choices such as MLPs, as shown in the appendix section \ref{mlp-predictor}. Even when using Transformers for both the skill predictor and the policy network, our compute requirement is comparable to the non-hierarchical Flat Transformer policy as we can get away with using fewer layers in each module.
% \sv{In our experiments, we use Transformer architectures with \LISA, but we also perform an experiment with a MLP-based skill predictor in the appendix. Even when using Transformers for both the skill predictor and the policy network, our compute requirement is comparable with that of the non-hierarchical Flat Transformer policy as we can get away with significantly smaller transformers for both the skill predictor and the BC policy.}

% \X policy uses a small horizon of past states $H=5,10$ as it only needs to predict actions for the current skill. Whereas works like Decison Transformer (DT)~\cite{decision-transformer} need to use \emph{all past observations} to condition on hierarchical language instructions. This is crucial for a task such as: "pick up the key and open the lock", where the first sub-goal has to be solved and be remembered as part of the context before attempting the next.  We find this issue in our flat baseline in ~\ref{sec:flat}.

\textbf{Language Encoder.} We use a pre-trained DistilBERT~\cite{distilbert} encoder to generate language embeddings from the text instruction. We fine-tune the language encoder end-to-end and use the full language embedding for each word token, and not a pooled representation of the whole text.

\textbf{Observation Encoder.} For image observations, we use convolution layers to generate embeddings. For simple state representations, we use MLPs.

\textbf{Skill Predictor.} 
The skill predictor network $f$ is implemented as a small Causal Transformer network that takes in the language embeddings and the observation embeddings at each time step. The language embeddings are concatenated at the beginning of the observation embeddings before being fed into the skill predictor. The network applies a causal mask hiding the future observations. 
% For a language instruction comprised of M word tokens, $l = (w_1, w_2, ..., w_M)$ and N observations $(s_{t-N}, ,.... s_t)$, we compute embeddings of size $[M, D]$ and $[N, D]$ respectively, where $D$ is embedding dimension. These embeddings are concatenated together to form a final embedding of size $[M+N, D]$.

\textbf{Policy Network.} Our policy network $\pi$, also implemented as a small Causal Transformer inspired by Decison Transformer (DT)~\cite{decision-transformer}. However, unlike DT, our policy is not conditioned on any reward signal, but on the skill code. The sequence length of $\pi$ is the horizon $H$ of the skills which is much smaller compared to the length of the full trajectory.

% \sv{repeated}
% The policy only sees a \emph{single} skill code, and past states over a small horizon $H=5, 10$. The skill code is resampled every $H$ steps, constraining the policy to execute the same skill over the horizon.

\textbf{Flat Decision Transformer Baseline.} Our flat baseline is based on DT and is implementation-wise similar to \X, but without a skill predictor network.  The policy here is a Causal Transformer, where we modify DT to condition on the language instruction embedding from a pre-trained DistillBERT text encoder instead of the future sum of returns. We found this baseline to be inefficient at handling long-range language instructions, needing sequence lengths of 1000 on complex environments such as BabyAI-BossLevel in our experiments. 

Since LISA has two transformers as opposed to just one in the flat baseline we ensured that the baseline and our method had a similar number of total parameters. To this end, the flat baseline uses Transformer network with 2 self-attention layers, and LISA’s skill predictor and policy use Transformer network with a single self-attention layer each. We also ensured that the embedding dimension and the number of heads in each layer were exactly the same in both LISA and the flat baseline. Details of this are provided in appendix sections \ref{sec:lisa_deets} and \ref{sec:baseline_deets} respectively. In fact, one could argue that LISA has less representation power because the policy transformer can only attend to the last H steps while the flat baseline can attend to the entire trajectory which is what makes it an extremely strong baseline. The flat baseline also uses the same pre-trained DistillBERT text encoder model as \X for dealing with natural language input.

% \sv{need to combine this with the points from the previous paragraphs}
% Here, the policy has an important difference to the \X policy network. Instead of seeing past observations over a small horizon window, it needs to be able to see all the past observations from the start of the trajectory. This is important as language can encode multiple sub-goals and in-lieu of using skill abstractions like \X, the policy has to use every observation from the start to figure out what to do next, reaching sequence lengths of 1000 on complex environments such as BabyAI-BossLevel.
% This is highly inefficient, needing 5x more compute, and almost double number of layers compared to \X.
% \sv{This compute discussion might have to be avoided}

\section{Experiments}
\label{experiment}

In this section, we evaluate \X on grid-world navigation and robotic manipulation tasks. We compare the performance of \X with a strong non-hierarchical baseline in the low-data regime. We then analyse our learnt skill abstractions in detail -- what they represent, how we can interpret them and how they improve performance on downstream composition tasks. 

For the sake of brevity, we present additional ablations in the Appendix~\ref{abl}, on doing manual planning with LISA skills (Section ~\ref{abl-plan}), transferring learned skills to different environments (Section~\ref{abl-transfer}) and learning continuous skills (Section~\ref{abl-cont}). 
\subsection{Datasets}

Several language-conditioned datasets have been curated as of late such as \cite{alfred, alfworld, babyai, lorl, calvin21, mattersim, chen2020touchdown, chen2021ask}. Nevertheless, a lot of these datasets focus on complex-state representations and navigation in 3D environments, making them challenging to train on and qualitatively analyze our skills as shown in Fig.~\ref{fig:word_cloud_gif}. We found BabyAI, a grid-world navigation environment and \lorl, a robotic manipulation environment as two diverse test beds that were very different from each other and conducive for hierarchical skill learning as well as detailed qualitative and quantitative analysis of our learned skills and we use them for our experiments.

\textbf{BabyAI Dataset.} The BabyAI dataset \cite{babyai} contains 19 levels of increasing difficulty where each level is set in a grid world and an agent sees a partially observed ego-centric view in a square of size 7x7. The agent must learn to perform various tasks of arbitrary difficulty such as moving objects between rooms, opening or closing doors, etc. all with a partially observed state and a language instruction. The language instructions for easy levels are quite simple but get exponentially more challenging for harder levels and contain several skills that the agent must complete in sequence (examples in appendix section \ref{babyai-examples}).
The dataset provides 1 million expert trajectories for each of the 19 levels, but we use $0.1-10\%$ of these trajectories to train our models. We evaluate our policy on a set of 100 different instructions from the gym environment for each level, which contain high percentage of unseen environments layouts and language instructions given the limited data we use for training. More details about this dataset can be found in Appendix~\ref{babyai-examples} and in the BabyAI paper.

\textbf{\lorl Sawyer Dataset.} This dataset~\cite{lorl} consists of \emph{pseudo-expert} trajectories or \textit{play data} collected from a replay buffer of a random RL policy and has been labeled with post-hoc crowd-sourced language instructions. Hence, the trajectories complete the language instruction provided but may not necessarily be optimal. 
Play data is inexpensive to collect \cite{lynch2019play} in the real world and it is important for algorithms to be robust to such datasets as well.
However, due to the randomness in the trajectories, this makes the dataset extremely difficult to use in a behavior cloning (BC) setting.
Despite this, we are able to achieve good performance on this benchmark and are able to learn some very useful skills. The \lorl Sawyer dataset contains 50k trajectories of length 20 on a simulated environment with a Sawyer robot. We evaluate on the same set of 6 tasks that the original paper does for our results in Table~\ref{tbl:perf}: \emph{close drawer, open drawer, turn faucet right, turn faucet left, move black mug right, move white mug down}.
We use two different settings -  with robot state space observations and partially-observed image observations.
More details can be found in the Appendixd \ref{lorl-examples} and in the \lorl paper. 

\subsection{Baselines}
\textbf{Original.} These refer to the baselines from the original paper for each dataset. For BabyAI, we trained their non-hierarchical RNN based method on different number of trajectories. Similarly, on \lorl we compare with the performance of language-conditioned BC. The original \lorl method uses a planning algorithm on a learned reward function to get around the sub-optimal nature of the trajectories. We found the BC baseline as a more fair comparison, as \LISA is trained using BC as well. Nonetheless, we compare with the original \lorl planner in Section~\ref{sec:composition} for composition tasks. \lorl results in Table~\ref{tbl:perf} refer to the performance on the 6 \textit{seen} instructions in the \lorl evaluation dataset, same as ones reported in the original paper.

\textbf{Flat Baseline.} We implement a non-hierarchical baseline using language-conditioned Decision Transformer denoted as \textbf{Lang DT}, the details of which are in section \ref{sec:flat}.

\subsection{How does performance of \X compare with non-hierarchical baselines in low-data regime?}
\label{sec:perf}
\looseness=-1
We consider three levels from the BabyAI environment and the \lorl Sawyer environment. For BabyAI, we consider the GoToSeq, SynthSeq and BossLevel tasks since they are challenging and require performing several sub-tasks one after the other. 
Since these levels contain instructions that are compositional in nature, when we train on limited data the algorithm must learn  skills which form complex instructions to generalize well to unseen instructions at test time.

Our experimental results are shown in Table~\ref{tbl:perf}. We train the models on a randomly sampled 1k, 10k and 100k trajectories from the full BabyAI dataset and 50k trajectories on the \lorl dataset. We use more data from the \lorl dataset because of the sub-optimal nature of the trajectories. 
\textbf{On all the environments, our method is competitive to or outperforms the strong non-hierarchical Decision Transformer baseline.}

The gap grows larger as we reduce the number of trajectories trained on, indicating that our method is able to leverage the common sub-task structures better and glean more information from limited data. 
As expected, with larger amounts of training data it becomes hard to beat the flat baseline since the model sees more compositions during training and can generalize better at test time~\cite{gpt3}.
As mentioned above, we evaluate on the same 6 \textit{seen} instructions the original \lorl paper did. We also evaluate the performance on varying language instructions on \lorl, similar to the original paper, with additional results in Appendix \ref{more-lorl-results}. 

We were pleasantly surprised that \textbf{\X is 2x better than the flat Lang-DT baseline on \lorl tasks}, reaching $40\%$ success rate using partial image observations despite the sub-optimal nature of the data. One explanation for this is that the discrete skill codes are able to capture \textit{different ways of doing the same task}, thereby allowing \X to \emph{learn an implicit multi-modal policy}.
This is not possible with the flat version as it has no way to compartmentalize these noisy trajectories, and perhaps tends to overfit on this noisy data, leading to performance degradation.

% \vspace{10pt}

\begin{table*}[t]
    \centering
	\small
% 	\vskip-3pt
% 	\tabcolsep 3pt
	\caption{\small \textbf{Imitation Results:} We show our success rates (in \%) compared to the original method and a flat non-hierarchical Decision Transformer baseline on each dataset over 3 seeds. \X outperforms all other methods in the low-data regime, and reaches similar performance as the number of demonstrations increases. Best method shown in \textbf{bold}.}  
	\label{tbl:perf}
	\vskip 5pt
	\begin{tabular}{l|c|c|c|c}
	   \toprule
		Task & Num Demos & Original & Lang DT & LISA \\ \midrule
	    BabyAI GoToSeq & $1k$ & $33.3 \pm 1.3$ & $49.3 \pm 0.7$  &  $\mathbf{59.4 \pm 0.9}$ \\
	    BabyAI GoToSeq & $10k$ & $40.4 \pm 1.2$ & $62.1 \pm 1.2$ & $\mathbf{65.4 \pm 1.6}$ \\
	    BabyAI GoToSeq & $100k$ & $47.1 \pm 1.1$ & $74.1 \pm 2.3$& $\mathbf{77.2 \pm 1.7}$ \\
	    \midrule
	    BabyAI SynthSeq & $1k$ & $12.9 \pm 1.2$ & $42.3 \pm 1.3$ & $\mathbf{46.3 \pm 1.2}$ \\
	    BabyAI SynthSeq & $10k$ & $32.6 \pm 2.5$  & $52.1 \pm 0.5$ & $\mathbf{53.3 \pm 0.7}$ \\
	    BabyAI SynthSeq & $100k$ & $40.4 \pm 3.3$ & $\mathbf{64.2 \pm 1.3}$ & $61.2 \pm 0.6$ \\
	    \midrule
		BabyAI BossLevel & $1k$ & $20.7 \pm 4.6$ & $44.5 \pm 3.3$ & $\mathbf{49.1 \pm 2.4}$  \\
	   % BabyAI BossLevel & $5k$ & $X$ &  & 100\% \\
	    BabyAI BossLevel & $10k$ & $28.9 \pm 1.3 $ & $\mathbf{60.1 \pm 5.5}$ & $58 \pm 4.1 $ \\
	    BabyAI BossLevel & $100k$ & $45.3 \pm 0.9 $ & $\mathbf{72.0 \pm 4.2}$ & $69.8 \pm 3.1 $ \\
	    \midrule
		\lorl \ - States (fully obs.) & $50k$ & $6 \pm 1.2$$^*$ & $33.3 \pm 5.6$ & $\mathbf{66.7 \pm 5.2}$ \\
		\lorl \ - Images (partial obs.) & $50k$ & $29.5 \pm 0.07$ & $15 \pm 3.4$ & $\mathbf{40 \pm 2.0}$ \\
		\hline
    \end{tabular}
    
{\raggedleft\scriptsize  $^*$ We optimized a language-conditioned BC model following the \lorl paper to the best of our abilities but could not get better performance.}

	\vskip-10pt
\end{table*}

\begin{figure}[h]
  \centering
  \vskip -10pt
  \includegraphics[width=\linewidth + 2cm]{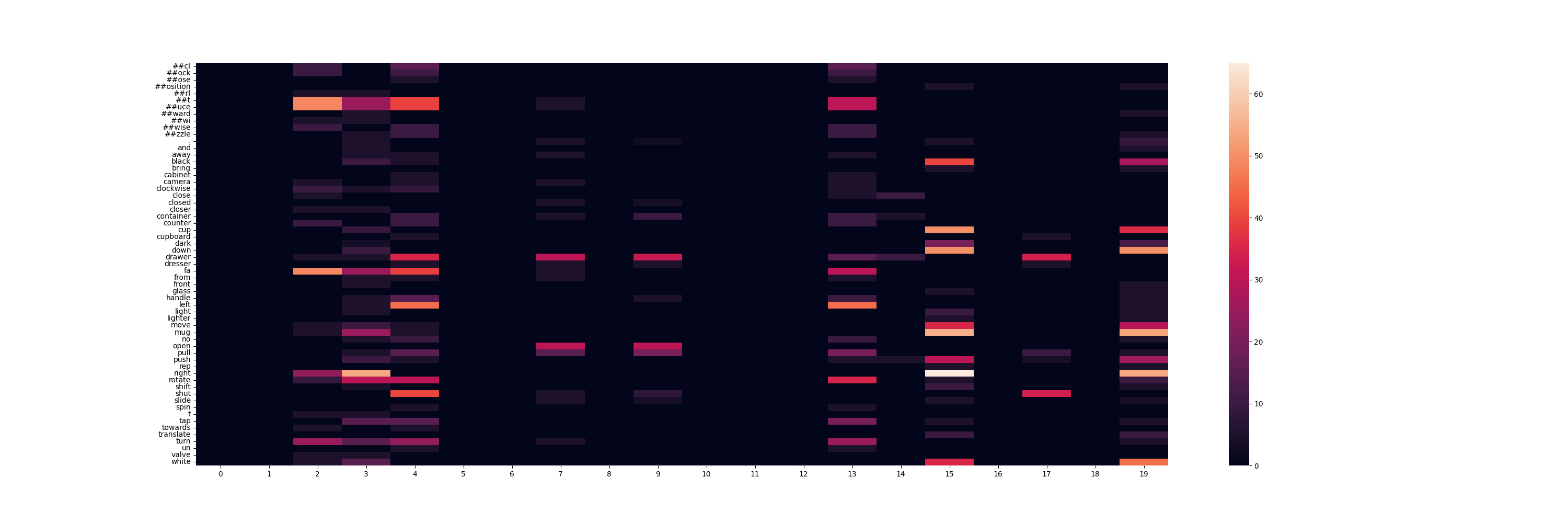}
   \vskip -10pt
  \caption{\small \textbf{\X Skill Heat map on \lorl.} We show the corresponding word frequency for 20 learned skills on \lorl (column normalized). The sparsity and the bright spots show that specific skills correspond to specific language tokens. (zoom in for best view)}
  \label{fig:heatmap_lorl}
    \vskip -8pt
\end{figure}

\subsection{What skills does \X learn? Are they diverse?}

% \begin{figure}[h]
%   \centering
%   \vskip -10pt
%   \includegraphics[width=\linewidth+1cm]{img/boss_word.pdf}
%   \vskip -10pt
%   \caption{\small \textbf{\X Skill Heat map on BabyAI BossLevel}}
%   \label{fig:heatmap_boss}
%     %  \vskip -10pt
% \end{figure}

\begin{wrapfigure}{r}{0.4\textwidth}
  \vskip -20pt
  %\centering
  \hspace*{-0.1cm}\includegraphics[width=\linewidth + 10pt]{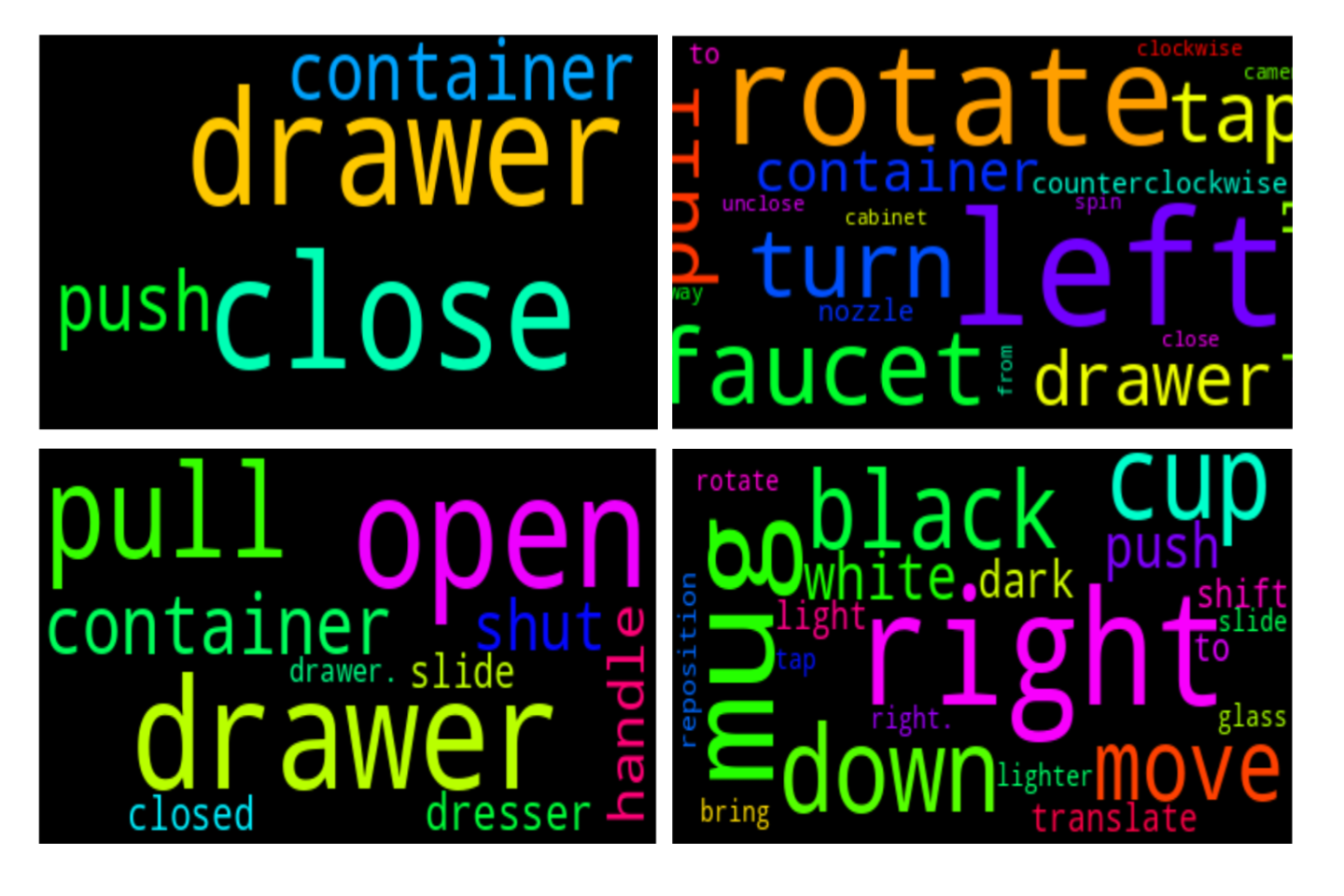}
  \vskip -5pt
  \caption{\small\textbf{Word clouds on \lorl}: We show the most correlated words for 4 different learnt skill codes on \lorl. We can see that the codes represent interpretable and distinguishable skills. For e.g, the code on the top left corresponds to closing the drawer. (note that container is a synonym for drawer in the \lorl dataset)}
  \label{fig:word_clouds}
  \vskip -15pt
\end{wrapfigure}

To answer this question, we analyse the skills produced by \X and the language tokens corresponding to each skill. We plot a heat map in Figure~\ref{fig:heatmap_lorl} corresponding to the correlation between the language tokens and skill codes. Here, we plot the map corresponding to the \lorl dataset. From the figure, we can see that certain skill codes correspond very strongly to certain language tokens and by extension, tasks. We also see the sparse nature of the heat maps which indicates that each skill corresponds to distinct language tokens. We also plot word clouds corresponding to four different options in the \lorl environment in Figure \ref{fig:word_clouds} and we notice that different options are triggered by different language tokens. From the figure, it is clear that the skill on the top left corner corresponds to \textit{close the drawer} and the skill on the top right corresponds to \textit{turn faucet left}.
Similar word clouds and heat maps for the BabyAI environments are in the appendix section \ref{babyai-wc}.

\subsection{Do the skills learned by \X correspond to interpretable behavior?}

We have seen that the different skills correspond to different language tokens, but do the policies conditioned on these skills behave according to the language tokens? To understand this, we fix the skill code for the entire trajectory and run the policy i.e. we are shutting off the skill predictor and always predicting the same skill for the entire trajectory. As we can see from the word cloud and the corresponding trajectory in Figure~\ref{fig:word_cloud_gif}, the behaviour for skill code 14 is exactly what we can infer from the language tokens in the word cloud -- \textit{close the drawer}. More such images and trajectories can be found in the appendix section \ref{more-gifs}.

\subsection{Why do \X learned skills show such a strong correlation to language?}
\label{sec:MI}
  % \vskip -20pt
\begin{wrapfigure}{r}{0.6\textwidth}
  \vskip -18pt
  \hspace*{-0.1cm}\includegraphics[width=\linewidth + 10pt]{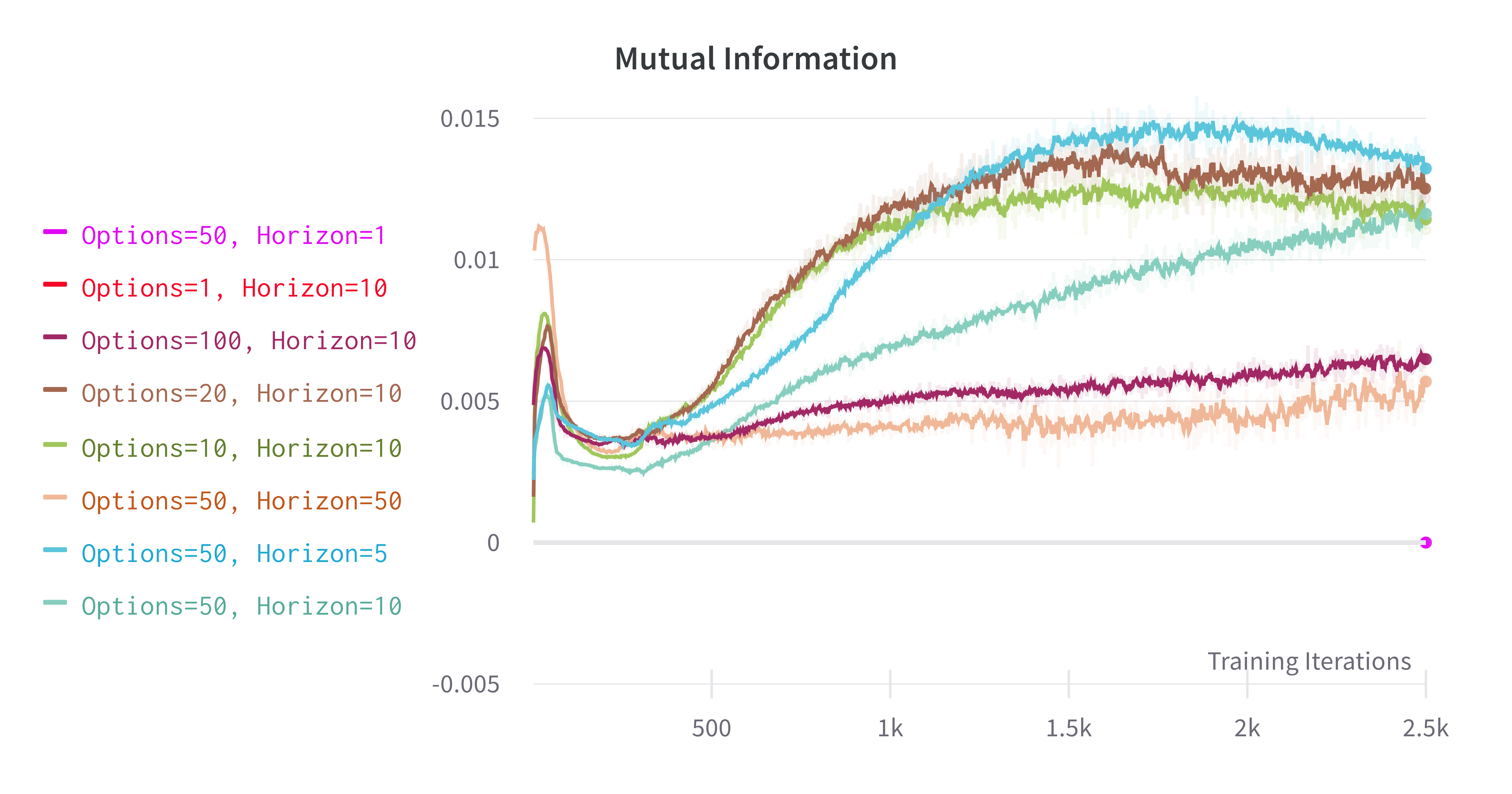} 
  \vskip -10pt
  \caption{\small\textbf{MI between language and skill codes:} We show the Mutual Information over training iterations for various settings of \X on the BabyAI BossLevel environment}
  \label{fig:mi}
  \vskip -30pt
\end{wrapfigure}

As mentioned in section \ref{sec:training}, the \emph{commitment loss} from VQ acts as a way to increase the MI between the language and the skill codes during training. This allows the codes to be highly correlated with language without any reconstruction losses. To analyze this, we plot the MI between the options and the language during training on the BabyAI BossLevel with 1k trajectories and the plot can be seen in Figure \ref{fig:mi}.
%When the number of skills is 1, $MI = 0$, as expected.
% We also notice that the MI is 0 when we change the option at every step i.e. when horizon is 1. 
% It makes sense as the skill predictor needs to choose about $70-80$ skills every episode and there are only 50 different skills to choose from so every skill corresponds to all language tokens over training. 
The plots show the MI increasing over training for a wide range of settings as we vary the number of skills and the horizon. In the ablation studies below, we report the success rate corresponding to each of these curves and we notice that there's almost a direct correlation with increasing MI and task performance. This is very encouraging since it clearly shows that the skills are encoding language and that directly impacts the performance of the behavior cloning policy.

\subsection{Can we use the learned skills to perform new composition tasks?}
\label{sec:composition}

\begin{wraptable}{R}{0.6\linewidth}
    \centering
	\small
	\vskip-18pt
	\tabcolsep 3pt
	\caption{\small \textbf{\X Composition Results:} We show our performance on the \lorl Sawyer environment on 15 unseen instructions compared to baselines}
	\label{tbl:comp}
	\begin{tabular}{l|c}
		Method & Success Rate (in \%) \\ \midrule
		 Flat & 13.33 $\pm$ 1.25 \\
		\lorl Planner &  18.18 $\pm$ 1.8 \\
		\X (Ours) & \textbf{20.89 $\pm$ 0.63} \\
		\hline
    \end{tabular}
	\vskip-10pt
\end{wraptable}

To test our composition performance, we evaluate on \lorl composition tasks using images in Table \ref{tbl:comp}. To this end, we handcraft 15 \textbf{unseen} composition instructions.
We have listed these instructions in the Appendix Table \ref{lorl-comp-instrs} with one such example \textit{``pull the handle and move black mug down''}. We ran 10 different runs of each instruction across 3 different seeds. As we can see, our performance is nearly 2x that of the non-hierarchical baseline. 
We also compare with the original \lorl planner on these composition tasks and we notice that we perform slightly better despite them having access to a reward function and a dynamics model pre-trained on 1M frames while \X is trained from scratch. We set the max number of episode steps to 40 from the usual 20 for all the methods while performing these experiments because of the compositional nature of the tasks.

\begin{wraptable}{R}{0.6\linewidth}
    \centering
	\small
	\vskip-18pt
	\tabcolsep 3pt
	\caption{\small \textbf{BabyAI:} \% of unseen instructions at test time for different training regimes.}
	\label{tbl:babyai-unseen}
	\begin{tabular}{l|c|c|c}
		 Environment & 1k trajs & 10k trajs & 100k trajs \\ \midrule
		 GoToSeq & 76\% & 63.8\% & 48\% \\
		 SynthSeq &  76.3\% & 66\% & 50.7\% \\
		 BossLevel & 77.3\% & 66.9\% & 51.3\% \\
		\hline
    \end{tabular}
	\vskip-10pt
\end{wraptable}

Note that results in Table \ref{tbl:perf} show compositionality performance on the BabyAI dataset as we train with 0.1\%-10\%  of the data. When we evaluate on the gym environment generating any possible language instruction from the BabyAI grammar, we may come across several unseen instructions at test time. 
To give a sense of the \% of unseen instructions for BabyAI when we evaluate on the gym environment, we take the different BabyAI environments and report the \% of unseen language instructions seen at test time for different training data regimes in Table \ref{tbl:babyai-unseen}. For each statistic, we sample 10,000 random instructions from the environment and check how many are unseen in the training dataset used, repeated over 3 different seeds.

\subsection{How does \X compare to simply doing K-Means clustering on the language and state embeddings?}
In \X, the VQ approach can be seen as taking the concatenated language-state inputs and projecting it into a learned embedding space.
VQ here simply learns $K$ embedding vectors that act as $K$ cluster centers for the projected input vectors in this embedding space, and allows for differentiability, enabling learning through backpropagation. 
This is also similar to propotypical methods used for few-shot learning and allows for deep differentiation clustering, giving an intuition of why it works.

To compare against $k$-means, we construct a simple unsupervised learning baseline that clusters trajectories in the training dataset using $k$-means.
Specifically, in the BabyAI BossLevel environment using 1k training trajectories, we take all concatenated language-state vectors for all trajectories in the dataset and cluster them using $k$-means and use the assigned cluster centers as the skill codes. 
We then learn a policy using these skill codes to measure their efficacy and found that \X has a performance of $\mathbf{49.1 \pm 2.4 \%}$ and $k$-means has a performance of $20.2 \pm 5.2 \%$ over 3 seeds.

Thus, we see that using the simple $k$-means skills is insufficient to learn a good policy to solve the BabyAI BossLevel task, as the skills are not representative enough of the language instructions. 
A reason for this is that language and state vectors lie in different embedding spaces, and K-means based on euclidean distance is not optimal on the concatenated vectors.

\section{Limitations and Future Work}
\label{future}

We present \X, a hierarchical imitation learning framework that can be used to learn interpretable skill abstractions from language-conditioned expert demonstrations. We showed that the skills are diverse and can be used to solve long-range language tasks and that our method outperforms a strong non-hierarchical baseline in the low-data regime. 

However, there are several limitations to \X and plenty of scope for future work. One limitation of \X is that there are several hyperparameters to tune that may affect performance like the number of options and the horizon for each option. It certainly helps to have a good idea of the task to decide these hyperparameters even though the ablations show that the method is fairly robust to these choices. Its also useful to learn the horizon for each skill by learning a termination condition and we leave this for future work. 

Although our method has been evaluated on the language-conditioned imitation learning setting, its not difficult to modify this method to make it work for image goals or demos, and in the RL setting as well. Its interesting to see if the vector quantization trick can be used to learn goal-conditioned skills in a more general framework.
 
%Maybe make the action DT a simple linear classifier?
%when language is not very useful for the task

\begin{ack}
We are thankful to John Schulman, Chelsea Finn, Karol Hausman and Dilip Arumugam for initial discussions regarding our method, and to Suraj Nair for providing help with the LOReL baseline.

This research was supported in part by NSF (\#1651565, \#1522054, \#1733686), ONR (N00014-19-1-2145),
AFOSR (FA9550-19-1-0024), FLI and Samsung.
\end{ack}

\small
\bibliographystyle{plainnat}
\bibliography{main}

\newpage
\appendix

\section*{Appendix}
\label{appendix}

% You can have as much text here as you want. The main body must be at most $8$ pages long.
% For the final version, one more page can be added.
% If you want, you can use an appendix like this one, even using the one-column format.

\section{Broader Societal Impact}
\label{soc_imp}
We introduce a new method for language-conditioned imitation learning to perform complex navigation and manipulation tasks.
Our intention is for this algorithm to be used in a real-world setting where humans can provide natural language instructions to robots that can carry them out.
However, we must ensure that the language commands that we provide to these agents must be well aligned with the objectives of humans and must ensure that we are aware of what actions the agent could take given said command.

\section{More visualizations}
\subsection{Generating heat maps and word clouds}

To generate heat maps and word clouds, for each evaluation instruction, we run the model and record all the skill codes used in the trajectory generated. We now tokenize the instruction and for each skill code used in the trajectory, record \textit{all} the tokens from the language instruction. Once we have this mapping from skills to tokens, we can plot heat maps and word clouds. This is the best we can do since we don't know exactly which tokens in the instruction correspond to the skills chosen. Therefore, these plots can tend to be a little noisy but we still see some clear patterns. Especially in BabyAI, since the vocabulary is small, we see that several skills correspond to the same tokens because many instructions contain the same tokens. But in \lorl because each task uses almost completely different words, we can see a very sparse heat map with clear correlations. 

For reader viewability and aiding the interpretability on the \LISA skills, we show below the unnormalized heatmaps showing the skill-word correlations, the column normalized heatmaps showing word frequencies for each skill as well as the row normalized showing the skill frequencies for each word.

\subsection{BabyAI}

\begin{figure}[h!]
  \centering
  %\fbox{\rule[-.5cm]{0cm}{4cm} \rule[-.5cm]{4cm}{0cm}}
  \includegraphics[width=50 em]{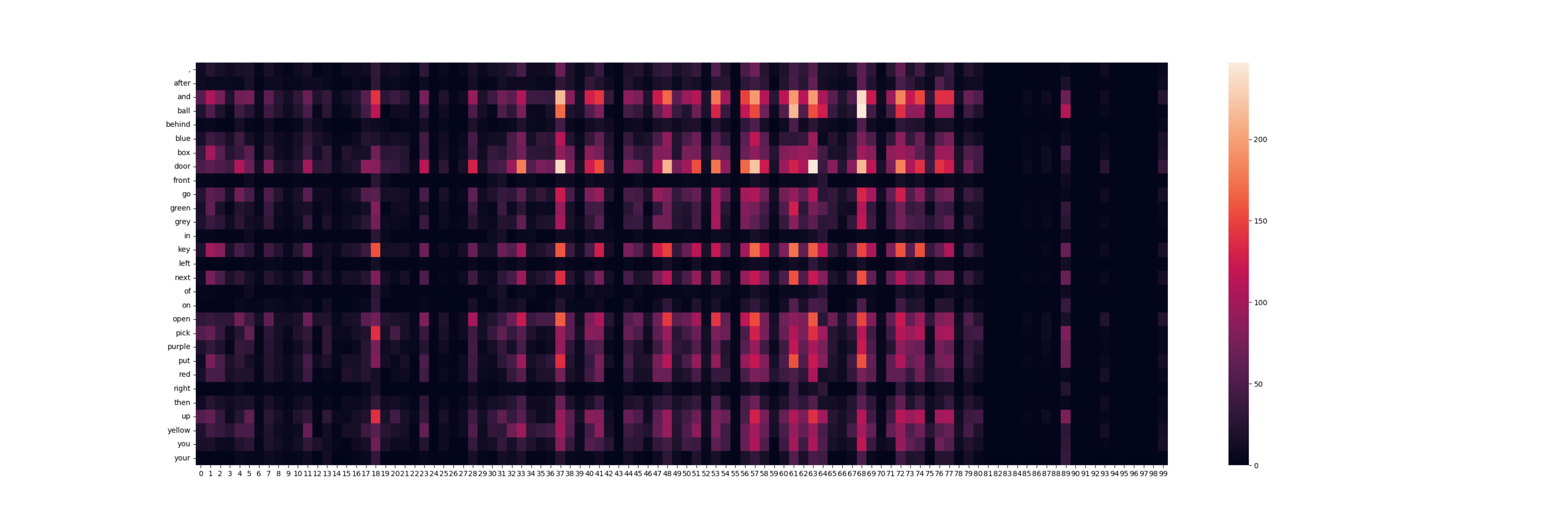}
  \caption{Skill Heat map on BabyAI BossLevel}
  \label{fig:boss_heat}
\end{figure}

\begin{figure}[h!]
  \centering
  %\fbox{\rule[-.5cm]{0cm}{4cm} \rule[-.5cm]{4cm}{0cm}}
  \includegraphics[width=50 em]{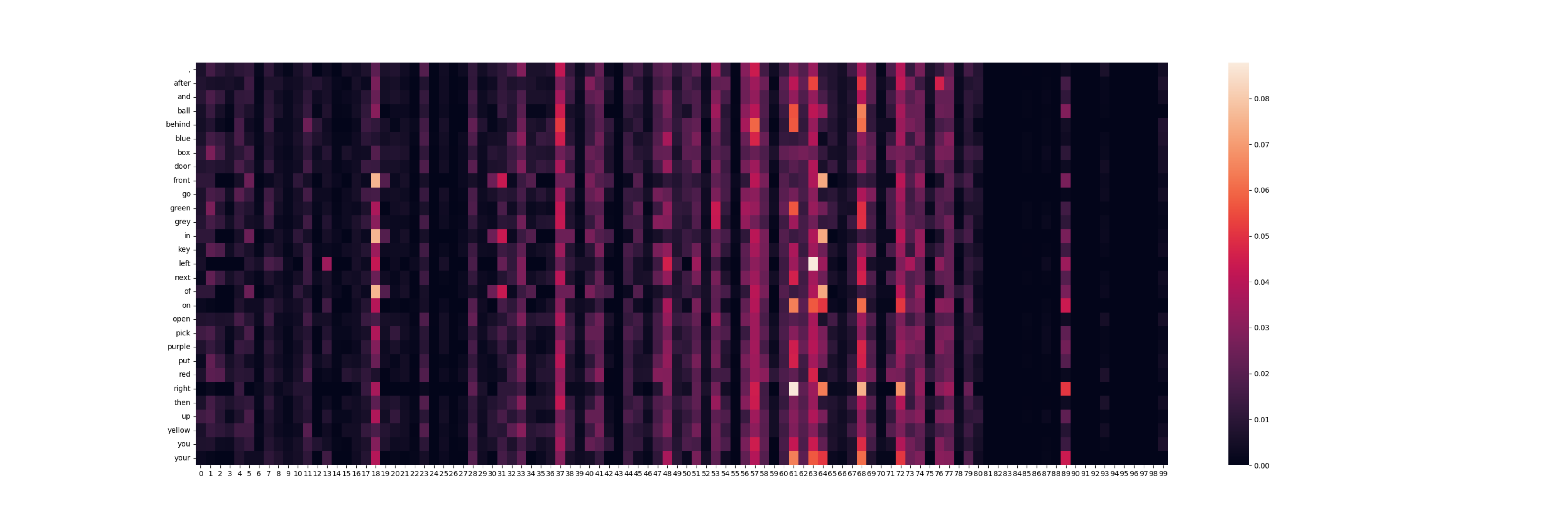}
  \caption{Word Freq. for each skill on BabyAI BossLevel (column normalized) }
  \label{fig:boss_heat2}
\end{figure}

\begin{figure}[h!]
  \centering
  %\fbox{\rule[-.5cm]{0cm}{4cm} \rule[-.5cm]{4cm}{0cm}}
  \includegraphics[width=50 em]{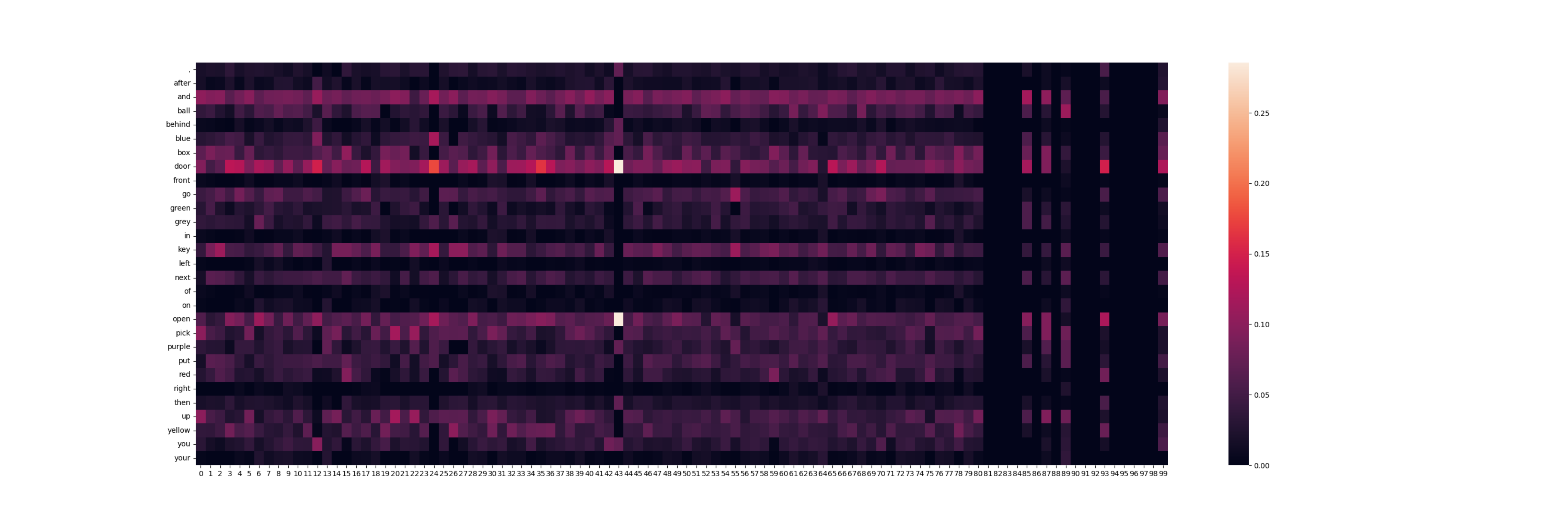}
  \caption{Skill Freq. for each word on BabyAI BossLevel (row normalized) }
  \label{fig:boss_heat3}
\end{figure}

\newpage 

\subsection{WordClouds}
\label{babyai-wc}

Due to the small vocabulary in BabyAI environment, its hard to generate clean word clouds, nevertheless, we hope they help with interpreting \LISA skills.

\begin{figure}[h!]
  \centering
  %\fbox{\rule[-.5cm]{0cm}{4cm} \rule[-.5cm]{4cm}{0cm}}
  \includegraphics[width=20 em]{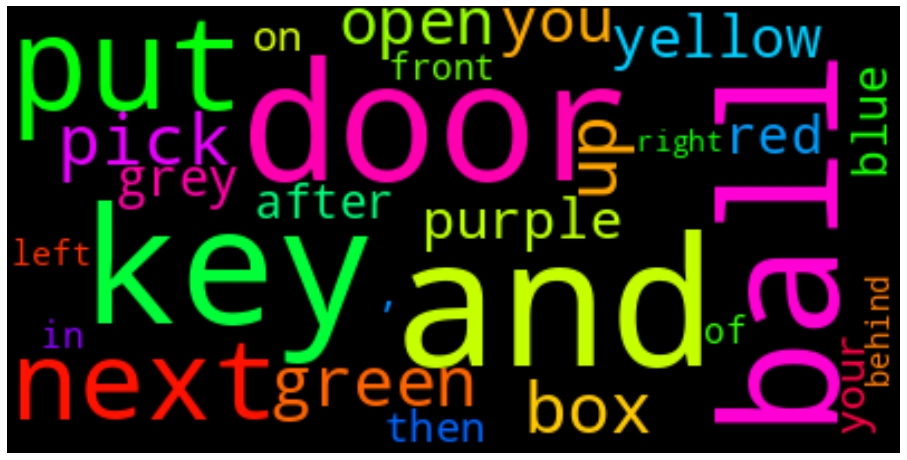}
  \caption{Word Cloud on BabyAI BossLevel for $z = 1$}
  \label{fig:boss_w1}
\end{figure}

\begin{figure}[h!]
  \centering
  %\fbox{\rule[-.5cm]{0cm}{4cm} \rule[-.5cm]{4cm}{0cm}}
  \includegraphics[width=20 em]{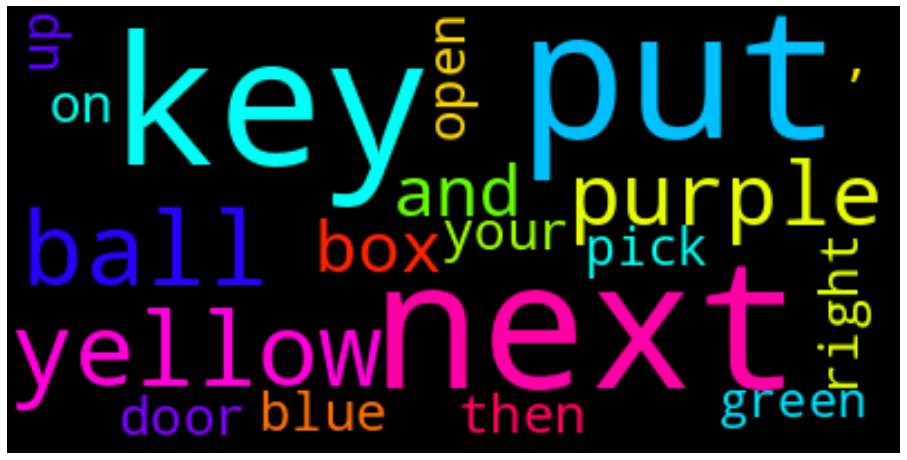}
  \caption{Word Cloud on BabyAI BossLevel for $z = 13$}
  \label{fig:boss_w2}
\end{figure}

\vskip - 10pt

\begin{figure}[h!]
  \centering
  %\fbox{\rule[-.5cm]{0cm}{4cm} \rule[-.5cm]{4cm}{0cm}}
  \includegraphics[width=20 em]{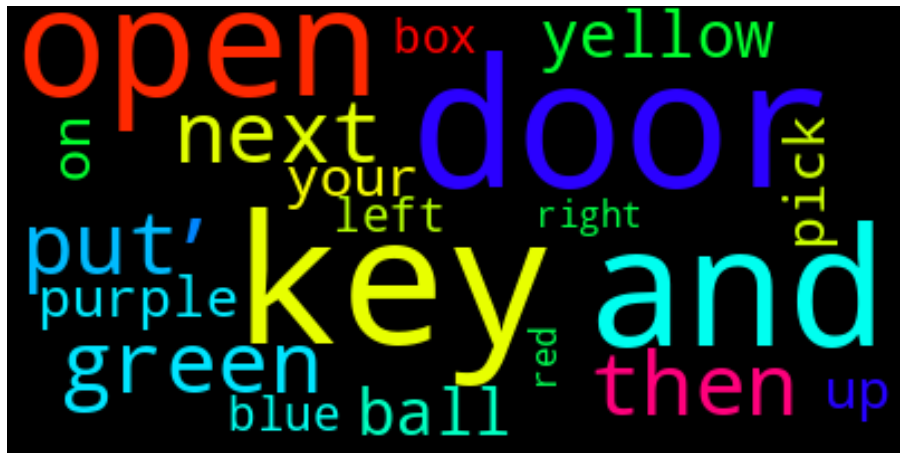}
  \caption{Word Cloud on BabyAI BossLevel for $z = 37$}
  \label{fig:boss_w3}
\end{figure}

\newpage

\subsection{\lorl Sawyer}

\begin{figure}[h!]
  \centering
  %\fbox{\rule[-.5cm]{0cm}{4cm} \rule[-.5cm]{4cm}{0cm}}
  \includegraphics[width=50 em]{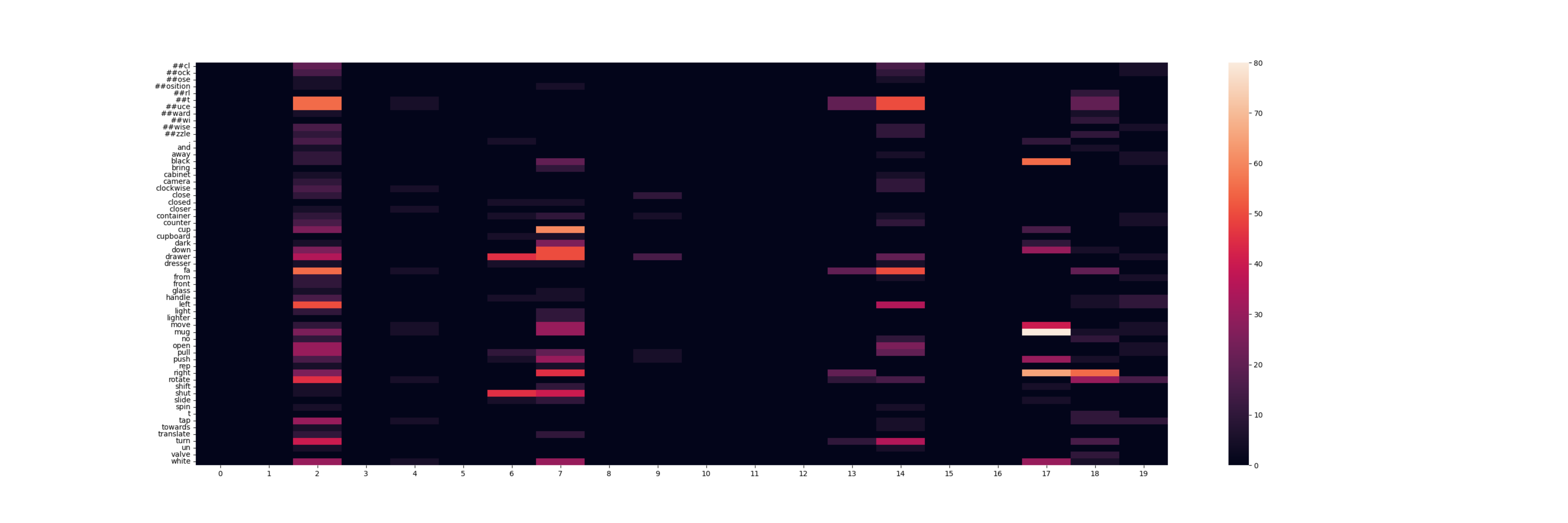}
  \caption{Skill Heat map on \lorl Sawyer}
  \label{fig:lorl_heat3}
\end{figure}

\begin{figure}[h!]
  \centering
  %\fbox{\rule[-.5cm]{0cm}{4cm} \rule[-.5cm]{4cm}{0cm}}
  \includegraphics[width=50 em]{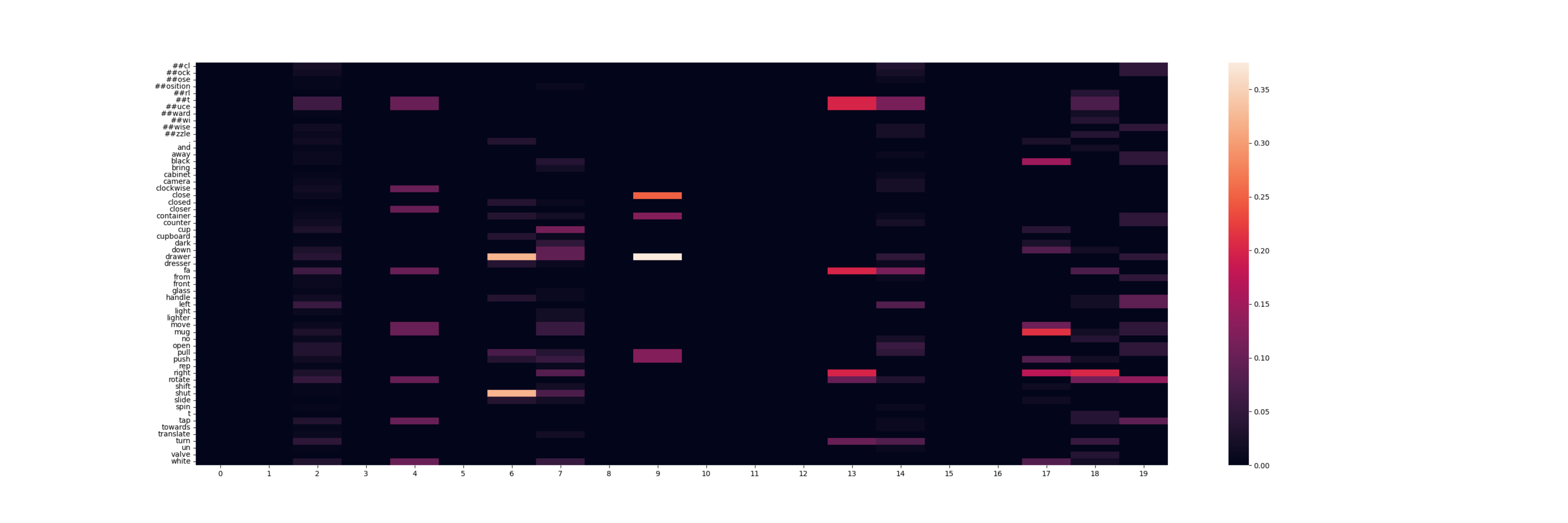}
  \caption{Word Freq. for each skill on \lorl Sawyer (column normalized) }
  \label{fig:boss_heat4}
\end{figure}

\begin{figure}[h!]
  \centering
  %\fbox{\rule[-.5cm]{0cm}{4cm} \rule[-.5cm]{4cm}{0cm}}
  \includegraphics[width=50 em]{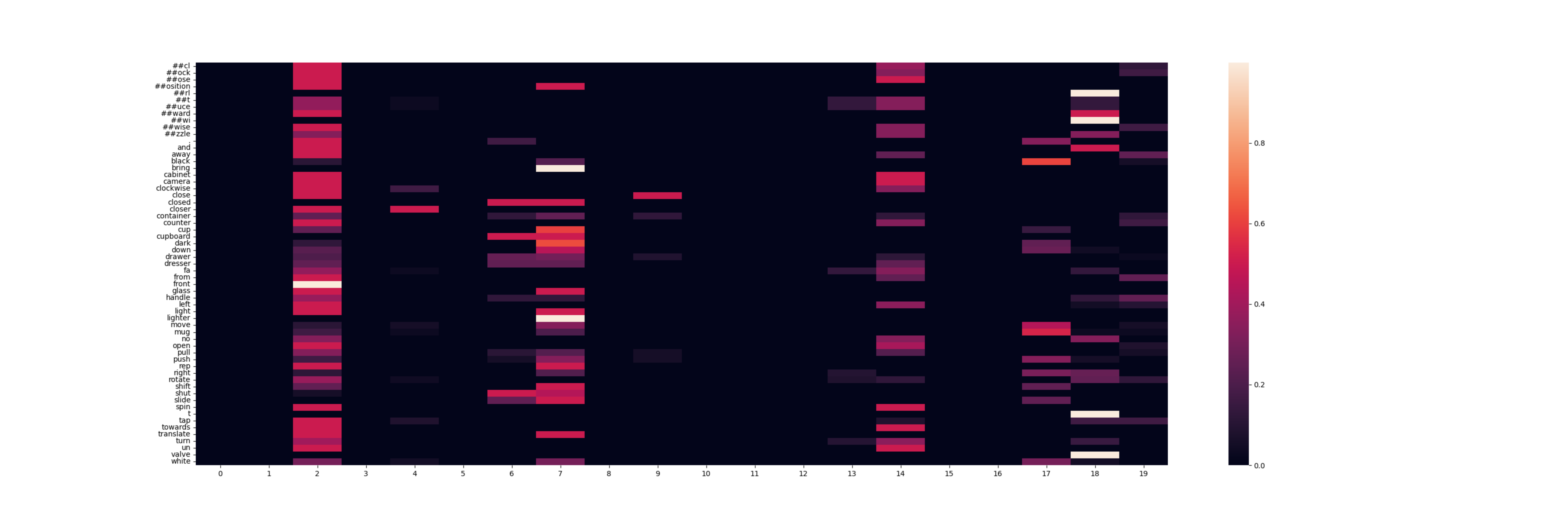}
  \caption{Skill Freq. for each word on \lorl Sawyer (row normalized) }
  \label{fig:boss_heat5}
\end{figure}

\newpage

\subsection{Behavior with fixed skills}
\label{more-gifs}
\begin{figure}[h!]
  \centering
  %\fbox{\rule[-.5cm]{0cm}{4cm} \rule[-.5cm]{4cm}{0cm}}
  \includegraphics[width=\linewidth]{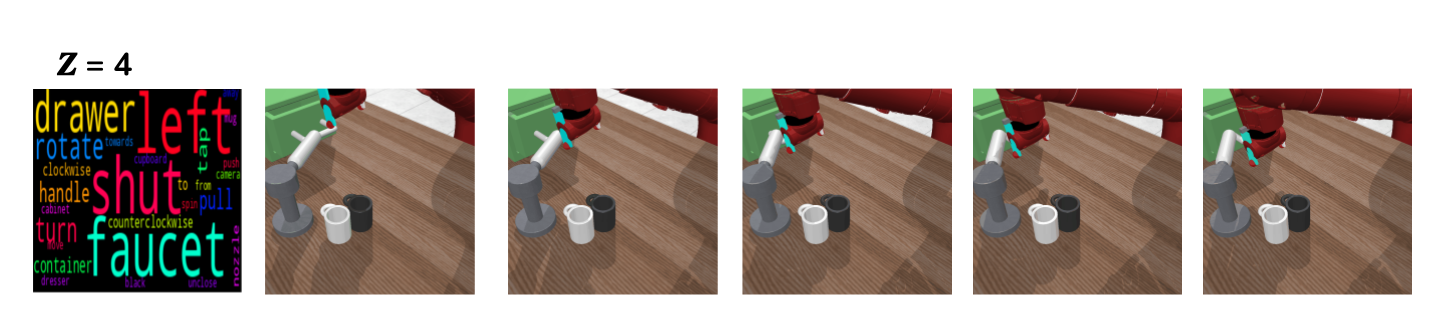}
  \caption{Behavior and language corresponding to skill code 4: ``turn faucet left''}
  \label{fig:word_cloud_4}
\end{figure}

\begin{figure}[h!]
  \centering
  %\fbox{\rule[-.5cm]{0cm}{4cm} \rule[-.5cm]{4cm}{0cm}}
  \includegraphics[width=\linewidth]{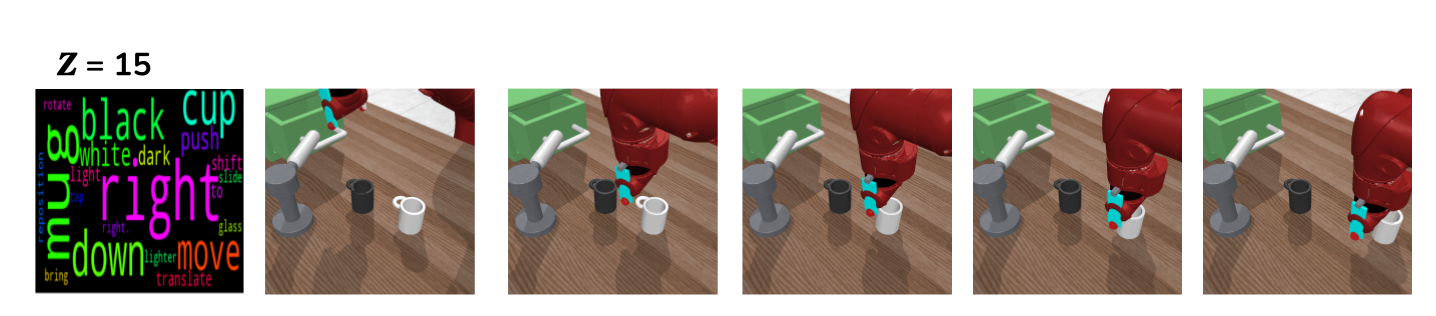}
  \caption{Behavior and language corresponding to skill code 15: ``move white mug right''}
  \label{fig:word_cloud_5}
\end{figure}

\begin{figure}[h!]
  \centering
  %\fbox{\rule[-.5cm]{0cm}{4cm} \rule[-.5cm]{4cm}{0cm}}
  \includegraphics[width=\linewidth]{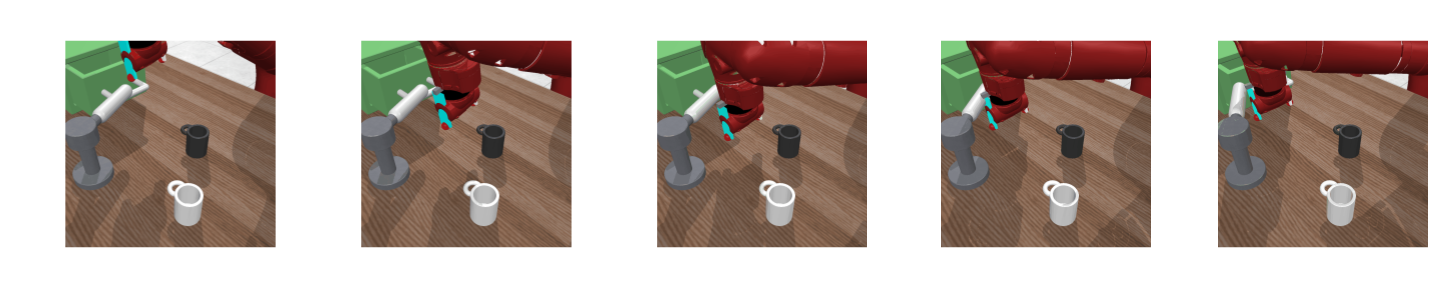}
  \caption{\lorl Composition task: ``close drawer and turn faucet left''}
  \label{fig:close_drawer_turn_faucet_left}
\end{figure}

\newpage

\section{Datasets}

\subsection{BabyAI Dataset}
\label{babyai-examples}

\begin{figure}[h!]
  \vskip -10 pt
  \centering
  %\fbox{\rule[-.5cm]{0cm}{4cm} \rule[-.5cm]{4cm}{0cm}}
  \includegraphics[width=18 em]{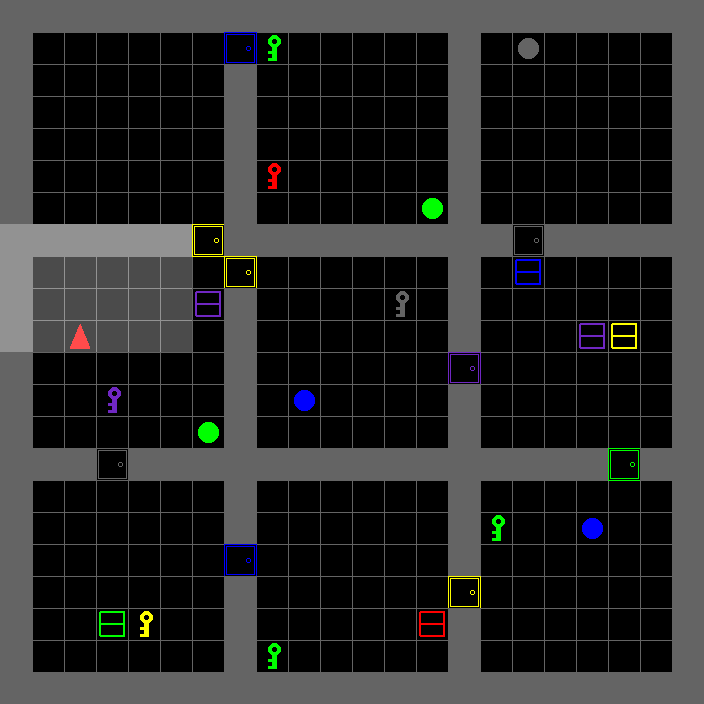}
  \vskip -5 pt
  \caption{BabyAI BossLevel}
  \label{fig:bosslevel}
\end{figure}

The BabyAI dataset \cite{babyai} contains 19 levels of increasing difficulty where each level is set in a grid world where an agent has a partially observed state of a square of side 7 around it. The agent must learn to perform various tasks of arbitrary difficulty such as moving objects between rooms, opening doors or closing them etc. all with a partially observed state and a language instruction. \\
Each level comes with 1 million language conditioned trajectories, and we use a small subset of these for our training. We evaluate our model on the environment provided with each level that generates a new language instruction and grid randomly. \\
We have provided details about the levels we evaluated on below. More details can be found in the original paper.

\subsubsection{GoToSeq}
Sequencing of go-to-object commands. \\
Example command: \textit{``go to a box and go to the purple door, then go to the grey door''} \\
Demo length: $72.7 \pm 52.2$

\subsubsection{SynthSeq}
Example command: \textit{``put a purple key next to the yellow key and put a purple ball next to the red box on your left after you put a blue key behind you next to a grey door''}\\
Demo length:  $81.8 \pm 61.3$

\subsubsection{BossLevel}
Example command: \textit{``pick up a key and pick up a purple key, then open a door and pick up the yellow ball''}\\
Demo length: $84.3 \pm 64.5$

\subsection{\lorl Sawyer Dataset}
\label{lorl-examples}
\begin{figure}[h!]
%   \vskip -5 pt
  \centering
  %\fbox{\rule[-.5cm]{0cm}{4cm} \rule[-.5cm]{4cm}{0cm}}
  \includegraphics[width=18 em]{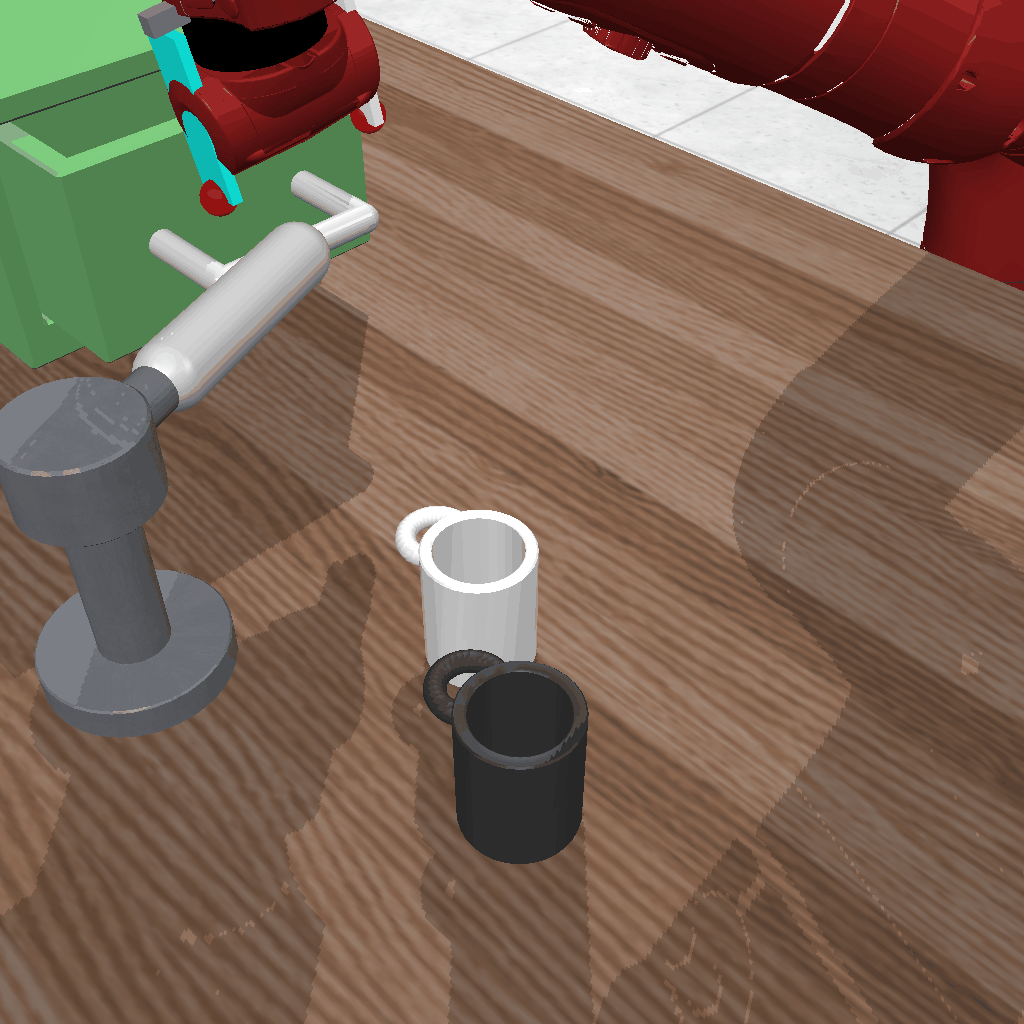}
  \vskip -5 pt
  \caption{\lorl Sawyer Environment}
  \label{fig:lorl_level}
\end{figure}

This dataset~\cite{lorl} consists of \emph{pseudo-expert} trajectories collected from a RL buffer of a a random policy and has been labeled with post-hoc crowd-sourced language instructions. Therefore, the trajectories complete the language instruction provided but may not necessarily be optimal. The Sawyer dataset contains 50k language conditioned trajectories on a simulated environment with a Sawyer robot of demo length 20. \\
We evaluate on the same set of instructions the original paper does for \ref{tbl:perf}, which can be found in the appendix of the original paper. These consist of the following 6 tasks and rephrasals of these tasks where they change only the noun, only the verb, both noun and verb and rewrite the entire task (human provided). This comes to a total of 77 instructions for all 6 tasks combined. An example is shown below and the full list of instructions can be found in the original paper. 

\begin{enumerate}
    \item Close drawer
    \item Open drawer
    \item Turn faucet left
    \item Turn faucet right
    \item Move black mug right
    \item Move white mug down
\end{enumerate}

\begin{table}[H]
    \centering
	\small
% 	\vskip-3pt
% 	\tabcolsep 3pt
	\caption{\small \lorl Example rephrasals for the instruction \textit{``close drawer''}}  
	\vskip-5pt
	\begin{tabular}{c|c|c|c|c}
		Seen & Unseen Verb & Unseen Noun & Unseen Verb + Noun & Human Provided \\ \midrule
		 close drawer & shut drawer & close container & shut container & push the drawer shut\\
		\hline
    \end{tabular}
	\vskip-10pt
\end{table}

For the composition instructions, we took these evaluation instructions from the original paper and combined them to form 12 new composition instructions as shown below. 

\begin{table}[H]
    \centering
	\small
% 	\vskip-3pt
% 	\tabcolsep 3pt
	\caption{\small \lorl Composition tasks}  
	\vskip-5pt
	\begin{tabular}{c}
	    \textbf{Instructions}\\
	    \hline
		open cabinet and move black mug right \\
        pull the handle and move black mug down \\
        shift white mug right \\
        shift black mug down \\
        shut drawer and turn faucet right \\
        close cabinet and turn faucet left \\
        turn faucet left and shift white mug down \\
        rotate faucet right and close drawer \\
        move white mug down and rotate faucet left \\
        open drawer and turn faucet counterclockwise \\
        slide the drawer closed and then shift white mug down \\
        rotate faucet left and move white mug down \\
        shift white mug down and shift black mug right \\
        turn faucet right and open cabinet \\
        move black mug right and turn faucet clockwise \\
		\hline
    \end{tabular}
    \label{lorl-comp-instrs}
	\vskip-10pt
\end{table}

We included the instructions \textit{``move white mug right''} and \textit{``move black mug down''} as composition tasks here in the hope that we may have skills corresponding to colors like black and white or directions like right and down that can be composed to form these instructions but we did not observe such behaviour. 

\section{Training details}
\label{sec:training_deets}
We plan to release our code on acceptance. Here we include all hyper-parameters we used. We implement our models in PyTorch. Our original flat baseline implementation borrows from Decision Transformer codebase which uses GPT2 to learn sequential behavior. However, we  decided to start from scratch in order to implement \LISA to make our code modular and easily support hierarchy. 
We use 1 layer Transformer networks for both the skill predictor and the policy network in our experiments for the main paper. We tried using large number of layers but found them to be too computationally expensive without significant performance improvements. In BabyAI and \lorl results we train all models for three seeds. 

For BossLevel environment we use 50 skill codes, for other environments we used the settings detailed in the table below:

\subsection{\X}
\label{sec:lisa_deets}

\begin{table}[H]
    \centering
	\small
% 	\vskip-3pt
% 	\tabcolsep 3pt
	\caption{\small \textbf{\LISA Hyperparameters}}
	\vskip-5pt
	\begin{tabular}{l}
$
\begin{array}{l}
\begin{array}{l|ll}
\text { Hyperparameter } & \text { BabyAI } & \text { LORL } \\
\hline \text { Transformer Layers } & 1 & 1 \\
\text { Transformer Embedding Dim } & 128 & 128 \\
\text { Transformer Heads } & 4 & 4 \\
\text { Skill Code Dim } & 16 & 16 \\
\text { Number of Skills } & 20 & 20 \\
\text { Dropout } & 0.1 & 0.1 \\
\text { Batch Size } & 128 & 128 \\
\text { Policy Learning Rate } & 1e-4 &  1e-4 \\
\text { Skill Predictor Learning Rate } & 1e-5 &  1e-5 \\
\text { Language Model Learning Rate } & 1e-6 &  1e-6 \\
\text { VQ Loss Weight } & 0.25 &  0.25 \\
\text { Horizon } & 10 & 10 \\
\text { VQ EMA Update } & 0.99 & 0.99 \\
\text { Optimizer } & \text { Adam } & \text { Adam } \\
\end{array}\\
\end{array}
$
    \end{tabular}
	\vskip-10pt
\end{table}

\subsection{Baselines}
\label{sec:baseline_deets}

\begin{table}[H]
    \centering
	\small
% 	\vskip-3pt
% 	\tabcolsep 3pt
	\caption{\small \textbf{Flat Baseline Hyperparameters}}
	\vskip-5pt
	\begin{tabular}{l}
$
\begin{array}{l}
\begin{array}{l|ll}
\text { Hyperparameter } & \text { BabyAI } & \text { LORL } \\
\hline \text { Transformer Layers } & 2 & 2 \\
\text { Transformer Embedding Dim } & 128 & 128 \\
\text { Transformer Heads } & 4 & 4 \\
\text { Dropout } & 0.1 & 0.1 \\
\text { Batch Size } & 128 & 128 \\
\text { Policy Learning Rate } & 1e-4 &  1e-4 \\
\text { Language Model Learning Rate } & 1e-6 &  1e-6 \\
\text { Optimizer } & \text { Adam } & \text { Adam } \\
\end{array}\\
\end{array}
$
    \end{tabular}
	\vskip-10pt
\end{table}

\textbf{Hyperparameter choices.} The hyperparameters were chosen approximately by us based on our estimate of the number of skills in the dataset and the complexity of the dataset. For example, we use 20 skill codes for \lorl experiments while we use 50 skill codes for the BabyAI BossLevel experiments as mentioned above. These hyperparameter choices are by no means exhaustive nor optimal as our ablation study in section \ref{sec:ablations} suggests that our choice of horizon is perhaps sub-optimal (we used H=10 for our experiments in Table \ref{tbl:perf} but appendix section \ref{sec:ablations} suggests that H=5 is better). The ablation study also suggests that the performance is fairly stable for a reasonable range of hyperparameter choices removing the burden from the practitioner.

\textbf{Baseline Implementations.} For the original baseline for BabyAI, we used the code from the original repository. For the \lorl baseline, we used the numbers from the paper for \lorl Images. For \lorl States BC baseline, we implemented it based on the appendix section of the paper. We ran the \lorl planner from the original repository for the composition instructions.

\subsection{Ablations}
As mentioned in the paper, all our ablations were performed on BabyAI BossLevel with 1k trajectories over a single seed for the sake of time. Unless otherwise specified, we use the following settings. We use a 1-layer, 4-head transformer for both the policy and the skill predictor. We use 50 options and a horizon of 10. We use a batch size of 128 and train for 2500 iterations. We use a learning rate of 1e-6 for the language model and 1e-4 for the other parameters of the model. We use 2500 warm-up steps for the DT policy. Training was done on Titan RTX GPUs. 

\subsection{MI Calculation}

We calculate mutual information between the language instructions ($L$) and the skill codes ($z$) by writing $MI(L, z) = H(L) - H(L|z)$. Our procedure for this is very simple and uses $\sim 10$ lines of code. Specifically, we first calculate $H(L|z)$  by assuming $p(L|z)$ to be gaussian $N(\mu, I)$ with unit variance, centered at the codebook vector $z$. We can calculate $\mu$ by finding the distance between the codebook vectors and the language vectors in the latent embedding space. We can similarly calculate the $H(L)$ by taking the expectation of $H(L|z)$ over all discrete codebook vectors.

\begin{lstlisting}[language=Python]
def MI(option_codes, lang_state_embeds):
    """Calculate entropy of options over each batch
    option_codes: [N, D]
    lang_state_embeds: [B, D]
    """
    with torch.no_grad():
        N, D = option_codes.shape
        lang_state_embeds = lang_state_embeds.reshape(-1, 1, D)

        embed = option_codes.t()
        flatten = rearrange(lang_state_embeds, '... d -> (...) d')

        distance = -(
            flatten.pow(2).sum(1, keepdim=True)
            - 2 * flatten @ embed
            + embed.pow(2).sum(0, keepdim=True)
        )
        cond_probs = torch.softmax(distance / 2, dim=1)

        # get marginal probabilities
        probs = cond_probs.mean(dim=0)
        entropy = (-torch.log2(probs) * probs).sum()

        # calculate conditional entropy with language
        # sum over options, and then take expectation over language
        cond_entropy = (-torch.log2(cond_probs) * cond_probs).sum(1).mean(0)
        return entropy - cond_entropy
\end{lstlisting}

\section{Detailed \lorl Sawyer results}
\label{more-lorl-results}
We provide details results on the \lorl evaluation instructions below for \X and the flat baseline in the same format as the original paper. The results are averaged over 10 runs. The time horizon used was 20 steps.

\begin{table}[h]
    \centering
	\small
% 	\vskip-3pt
% 	\tabcolsep 3pt
	\caption{\small \textbf{Task-wise success rates (in \%) on \lorl Sawyer.}}  
	\vskip-5pt
	\begin{tabular}{l|c|c}
		Task & Flat & \X \\ \midrule
		close drawer &  10 & \textbf{100}\\
		open drawer &  \textbf{60} & 20\\
		turn faucet left & 0 & 0\\
		turn faucet right & 0 & \textbf{30}\\
		move black mug right & 20 & \textbf{60}\\
		move white mug down & 0 & \textbf{30}\\
		\hline
    \end{tabular}
% 	\vskip-10pt
\end{table}

\begin{table}[h]
    \centering
	\small
% 	\vskip-3pt
% 	\tabcolsep 3pt
	\caption{\small \textbf{Rephrasal-wise success rates (in \%) on \lorl Sawyer.}}  
	\vskip-5pt
	\begin{tabular}{l|c|c}
		Rephrasal Type & Flat & \X \\ \midrule
		seen &  15 & \textbf{40}\\
		unseen noun &  13.33 & \textbf{33.33}\\
		unseen verb & 28.33 & \textbf{30}\\
		unseen noun+verb & 6.7 & \textbf{20}\\
		human & 26.98 & \textbf{27.35}\\
		\hline
    \end{tabular}
	\vskip-10pt
\end{table}

\section{More experiments}
\label{abl}

\subsection{Ablation Studies}
\label{sec:ablations}
For the sake of time, all our ablations were performed with a 1-layer, 4-head transformer for the skill predictor and for the policy. All our ablations are on the BabyAI-BossLevel environment with 1k expert trajectories.

Our first experiment varies the horizon of the skills. The table below shows the results on BabyAI BossLevel for 4 different values of the horizon. We see that the method is fairly robust to the different choices of horizon unless we choose a very small horizon. For this case, we notice that a horizon of 5 performs best, but this could vary with different tasks.

\begin{table}[H]
    \centering
	\small
	\vskip-3pt
	\tabcolsep 3pt
	\caption{\small \textbf{Ablation on horizon.} We fixed the number of options to be 50 for these experiments}  
	\label{tbl:abl_perf}
	\begin{tabular}{l|c|c|c|c}
	\toprule
		Horizon & 1 & 5 & 10 & 50 \\
		\midrule
		Success Rate (in \%) & 32 & \textbf{52} & 47 & 47\\ 
		\hline
    \end{tabular}
	\vskip-10pt
\end{table}

We also tried varying the number of skills the skill-predictor can choose from and found that this hyperparameter is fairly robust as well unless we choose an extremely high or low value. We suspect using more skills worsens performance because it leads to a harder optimization problem and the options don't clearly correspond to specific language skills. 
%Using only one skill also doesn't work very well because its hard to effectively capture the space of all language instructions with just one skill.

\begin{table}[H]
    \centering
	\small
	\vskip-3pt
	\tabcolsep 3pt
	\caption{\small \textbf{Ablation on number of options.} We fixed the horizon to be 10 for these experiments} 
	\label{tbl:ablations}
	\begin{tabular}{l|c|c|c|c|c}
	\toprule
% 		Number of Options & 1 & 10 & 20 & 50 & 100 \\
% 		\midrule
% 		Success Rate (in \%) & 44 & \textbf{47} & \textbf{47} & \textbf{47} & 43\\ 
        Number of Options & 10 & 20 & 50 & 100 \\
 		\midrule
 		Success Rate (in \%) & \textbf{47} & \textbf{47} & \textbf{47} & 43\\ 
		\hline
    \end{tabular}
	\vskip-10pt
\end{table}

\subsection{Can we leverage the interpretability of the skills produced by \X for manual planning?}
\label{abl-plan}

\begin{figure}[h!]
  \vskip -5 pt
  \centering
  %\fbox{\rule[-.5cm]{0cm}{4cm} \rule[-.5cm]{4cm}{0cm}}
  \includegraphics[width=\linewidth]{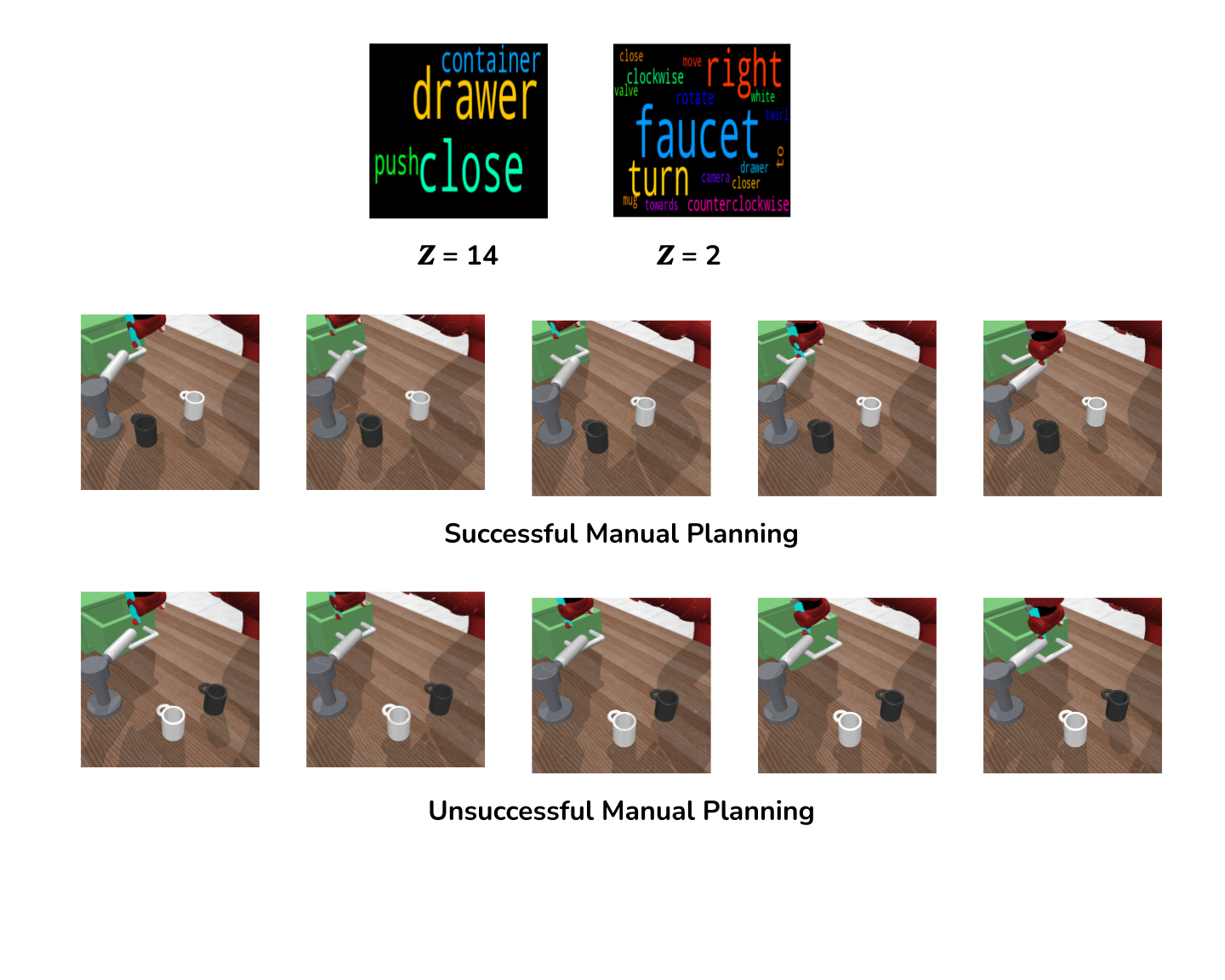}
  \vskip -50 pt
  \caption{\small We show a successful manual planning and unsuccessful manual planning example for the instruction ``close the drawer and turn the faucet right''}
  \label{fig:manual_planning}
\end{figure}

Since our skills are so distinct and interpretable, its tempting to try and manually plan over the skills based on the language tokens they encode. In the \lorl environment, using the same composition tasks as section above, we observe the word clouds of options and simply plan by running the fixed option code corresponding to a task for a certain horizon and then switch to the next option corresponding to the next task. This means we are using a manual (human) skill predictor as opposed to our trained skill predictor. While this doesn't work as well because skills are a function of both language and trajectory and we can only interpret the language part as humans, it still shows how interpretable our skills are as humans can simply observe the language tokens and plan over them to complete tasks. We show a successful example and a failure case below for the instruction \textit{``close drawer and turn faucet right''} in figure \ref{fig:manual_planning}. We first observe that the two skills we want to compose are $Z=14$ and $Z=2$ as shown by the word clouds. We then run skill 14 for 20 steps and skill 2 for 20 steps. In the failure case, the agent closes the drawer but then pulls it open again when trying to turn the faucet to the right.

\subsection{Do the skills learned transfer effectively to similar tasks?}
\label{abl-transfer}

\begin{figure}[h!]
  \vskip -5 pt
  \centering
  %\fbox{\rule[-.5cm]{0cm}{4cm} \rule[-.5cm]{4cm}{0cm}}
  \includegraphics[width=\linewidth]{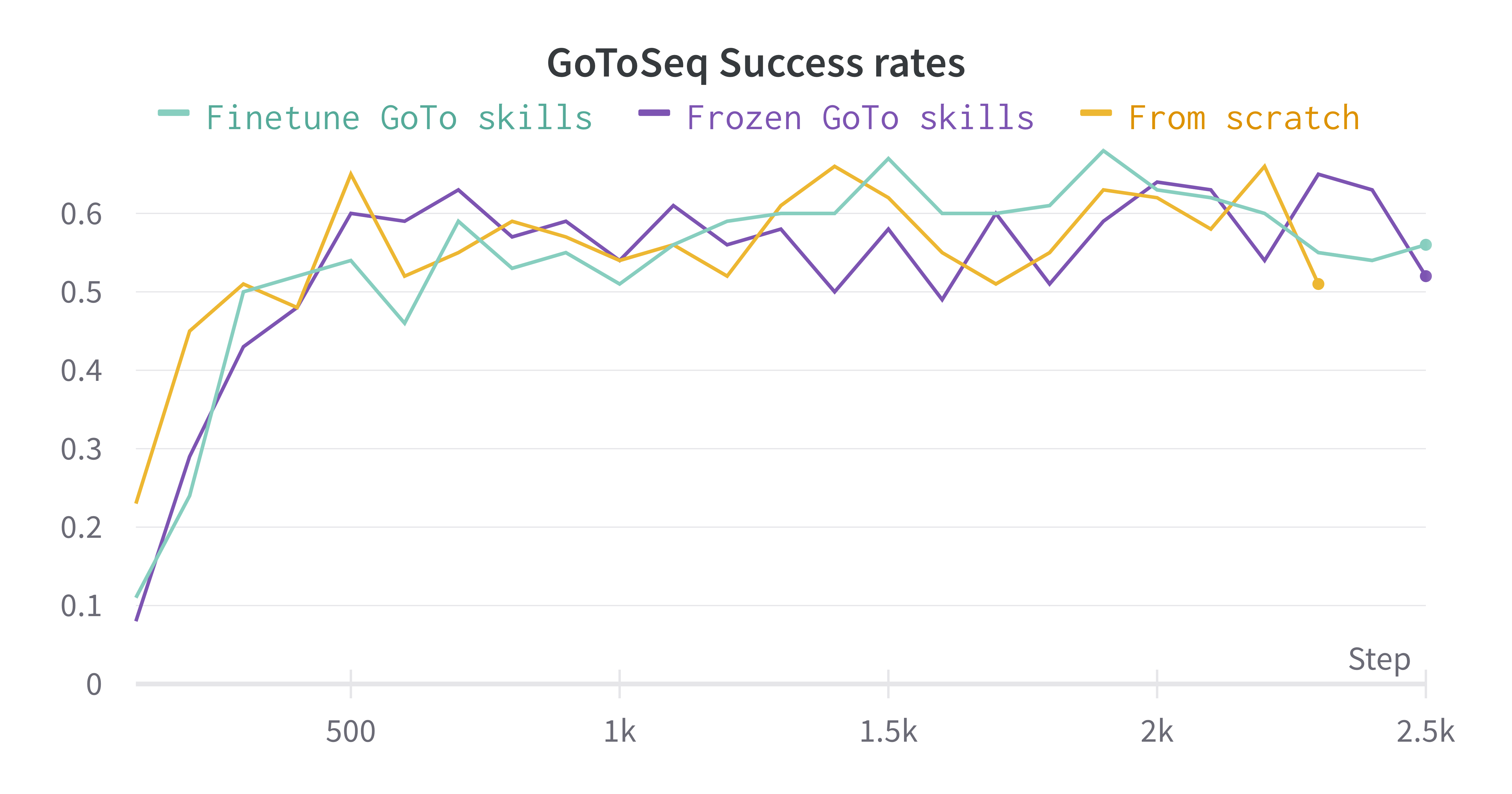}
  \vskip -10 pt
  \caption{\small Transferring skills on the BabyAI GoToSeq environment}
  \label{fig:gotoseq_transfer}
\end{figure}

\begin{figure}[h!]
  \vskip -5 pt
  \centering
  %\fbox{\rule[-.5cm]{0cm}{4cm} \rule[-.5cm]{4cm}{0cm}}
  \includegraphics[width=\linewidth]{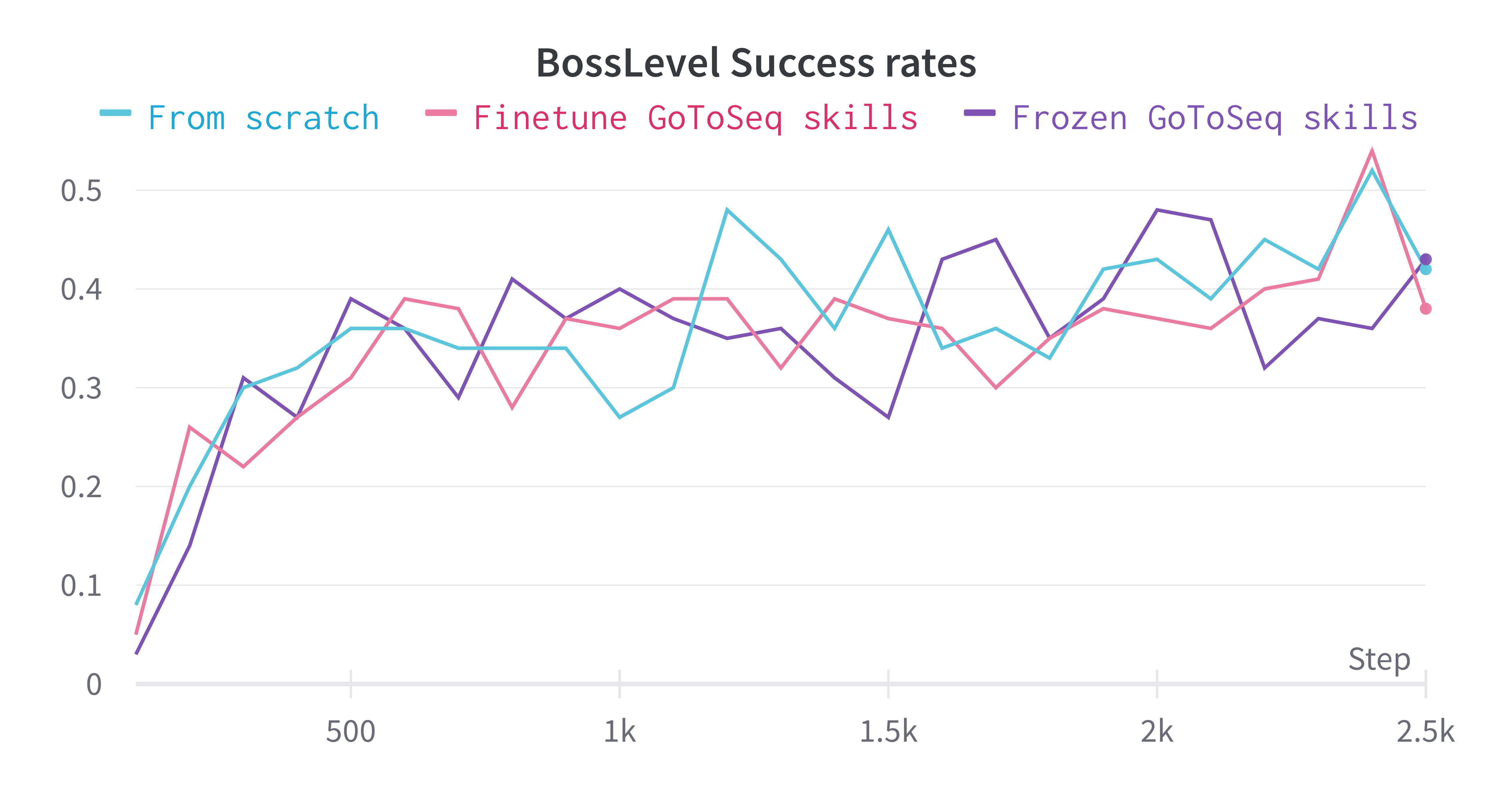}
  \vskip -10 pt
  \caption{\small Transferring skills on the BabyAI BossLevel environment}
  \label{fig:bosslevel_transfer}
\end{figure}

We want to ask the question whether we can use the skills learned on one task as a initialization point for a similar task or even freeze the learned skills for the new task. To this end, we set up experiments where we trained \X on the BabyAI GoTo task with 1k trajectories and tried to transfer the learned options to the GoToSeq task with 1k trajectories. Similarly we trained \X on GoToSeq with 1k trajectories and tried to transfer to BossLevel with 1k trajectories. The results are shown in figures \ref{fig:gotoseq_transfer} and \ref{fig:bosslevel_transfer} respectively.

As we can see from the GoToSeq experiment in figure \ref{fig:gotoseq_transfer}, there is no major difference between the three methods. We notice that we can achieve good performance even by holding the learned option codes from GoTo frozen. This is because the skills in GoTo and GoToSeq are very similar except that GoToSeq composes these skills as tasks. We also notice that finetuning doesn't make a big difference --  once again probably because the skills are similar for both environments. \\
In the BossLevel case in \ref{fig:bosslevel_transfer}, however, we do notice that the frozen skills perform slightly worse than the other two methods. This is because the BossLevel contains more skills than those from GoToSeq. We also notice that the performance of finetuning and starting from scratch is nearly the same. This could be because the meta-controller needs to adapt to use the new skills in the BossLevel environment anyway and there is no benefit from loading learned options here.

\subsection{Are the skill codes in \X chosen because of environment affordances and not language?}
A natural hypothesis for \X choosing relevant skill codes for each task could be that they are chosen because of the environment affordances and not because of the language instructions.
For example, \X chooses the code corresponding to opening a drawer because the drawer is closed in the environment or because the robot arm is close to it when initialized. We systematically show that this is not the case.
We initialize the \lorl environment with the drawer wide open and test all possible combinations of words meaning the same as ``drawer" and ``close".
\begin{itemize}
    \item We issue the following language commands: ``close drawer”, ``close container”, ``push drawer” and ``push container”. \X is unable to solve the task ``push container” but for the other three tasks, over 20 different environment seeds, \X generates the skill code 14 (as shown in Figure \ref{fig:word_cloud_gif}), all 20 times for each instruction.
    \item We then issue the commands containing the object ``drawer" with actions meaning different than ``close": ``rotate drawer”, ``rotate container”, ``move container down” and ``move drawer down”, ``open drawer", ``open container", ``pull drawer", ``pull container". We observe that none of these instructions used skill code 14 even once out of 20 different runs of each instruction.
    \item Next, we ran the commands containing the object different from ``drawer" and actions same as ``close": ``close mug”, ``close faucet”, ``push mug” and ``push faucet” and none of these instructions use the skill code 14 even once out of 20 different runs of each instruction.
\end{itemize}

These experiments indicate that the skill code 14 clearly is generated by \X only when the ``close drawer" instruction or a synonym is given in the language instruction. This clearly highlights that the codes are highly correlated to language and are not just being chosen based on the affordances of the environment.

\subsection{State-based skill-predictor}
\label{mlp-predictor}
\begin{figure}[h!]
  \centering
  %\fbox{\rule[-.5cm]{0cm}{4cm} \rule[-.5cm]{4cm}{0cm}}
  \includegraphics[width=45 em]{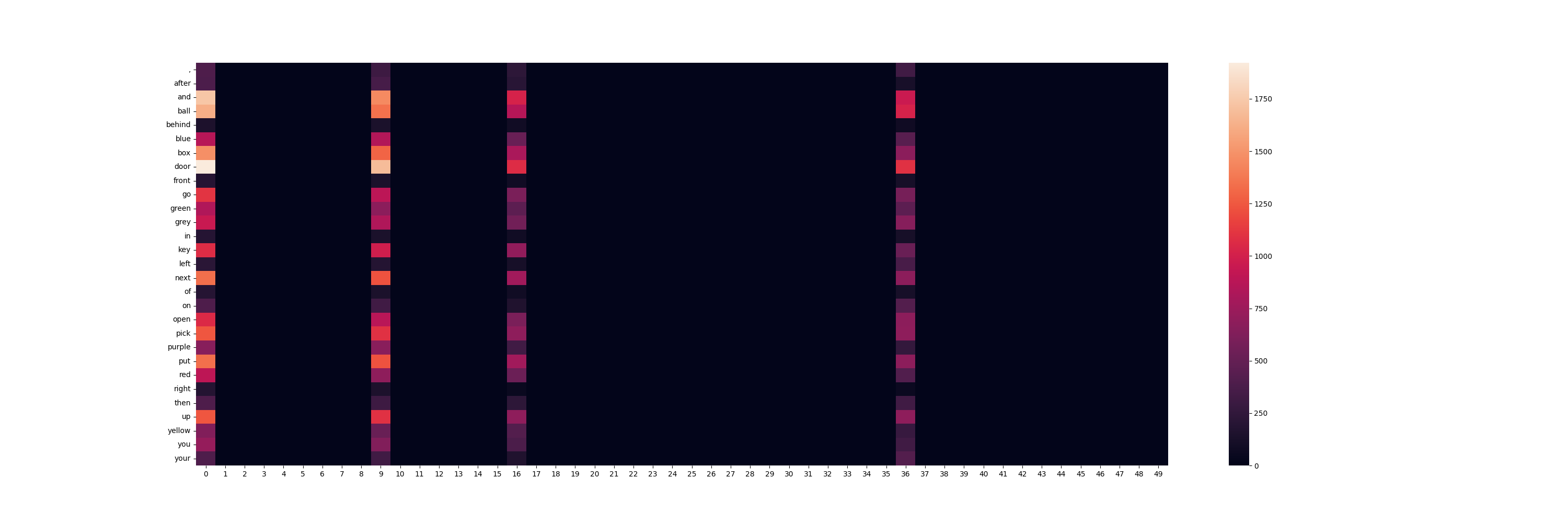}
  \caption{State-based skill predictor heat map shows that the model tends to use fewer options compared to figure \ref{fig:boss_heat}}
  \label{fig:option_heatmap}
\end{figure}

We have already spoken in section \ref{sec:training} about the fact that our method using two transformers doesn't necessarily mean its more compute-heavy than the non-hierarchical counterpart. But to test whether we really need two transformers, we perform an ablation study that replaces the skill predictor to be just a state-based selector MLP as opposed to a trajectory-based transformer. Our results show that the performance is only slightly worse when using a state-based skill predictor in this case, but once again this may not be the case in more complex environments. However, we notice that the skills collapsed in this case and the model tends to use fewer skills than normal as shown in figure \ref{fig:option_heatmap}. This is expected because the skill predictor is now predicting skills with much less information.

\begin{table}[h]
    \centering
	\small
% 	\vskip-3pt
% 	\tabcolsep 3pt
	\caption{\small \textbf{Comparing state-based MLP skill predictor vs trajectory-based transformer skill predictor.} We fixed the number of options to be 50 and horizon as 10 for these experiments}  
	\vskip-5pt
	\begin{tabular}{l|c}
		Skill Predictor Architecture & Success Rate \\ \midrule
		State-based MLP & 46\% \\
		Trajectory-based Transformer & \textbf{47\%} \\
		\hline
    \end{tabular}
	\vskip-10pt
\end{table}

\subsection{Continuous skill codes}
\label{abl-cont}
We also compare to the non-quantized counterpart where we learn skills from a continuous distribution as opposed to a categorical distribution. We expect this to perform better because the skill predictor has access to a larger number of skill codes to choose from and this is what we observe in table \ref{tbl:quant1}. However, this comes at the price of interpretability and its harder to interpret and choose continuous skill codes than discrete codes. 
We also observe that on the \lorl with states environment, using discrete codes performs better than using continuous codes (table \ref{tbl:quant2}). This could be because learning a multi-modal policy with discrete skills is an easier optimization problem than learning one with continuous skills (see the end of section \ref{sec:perf}).

\begin{table}[H]
    \centering
	\small
% 	\vskip-3pt
% 	\tabcolsep 3pt
	\caption{\small \textbf{Ablation on Quantization on BabyAI BossLevel.} We fix number of options to 50 and  horizon to 10}  
	\vskip-5pt
	\label{tbl:quant1}
	\begin{tabular}{l|c}
		Skill codes & Success Rate (in \%) \\ \midrule

		Continuous & \textbf{51} \\
		Discrete & 47 \\
		\hline
    \end{tabular}
	\vskip-10pt
\end{table}

\begin{table}[H]
    \centering
	\small
% 	\vskip-3pt
% 	\tabcolsep 3pt
	\caption{\small \textbf{Ablation on Quantization on \lorl with states on seen tasks.} We fix number of options to 20 and  horizon to 10}  
	\vskip-5pt
	\label{tbl:quant2}
	\begin{tabular}{l|c}
		Skill codes & Success Rate (in \%) \\ \midrule

		Continuous & 60.0 \\
		Discrete & \textbf{66.7} \\
		\hline
    \end{tabular}
	\vskip-10pt
\end{table}

% \subsection{Fine-tune vs Fixed LM}
% We perform an ablation study to understand the impact of finetuning the DistilBert language model \cite{distilbert} vs keeping it fixed. We conducted the experiment on the BabyAI BossLevel with the same setting as the other ablation studies. The performance difference is shown in the table below.

% \begin{table}[H]
%     \centering
% 	\small
% % 	\vskip-3pt
% % 	\tabcolsep 3pt
% 	\caption{\small Effect of finetuning the language model}  
% 	\vskip-5pt
% 	\begin{tabular}{c|c}
% 	    Language Model & Success Rate \\
% 		\hline
% 		Finetuned & 47\% \\
% 		Frozen & 47\% \\
% 		\hline
%     \end{tabular}
% 	\vskip-10pt
% \end{table}

% We find that there is no difference in performance between finetuning and freezing the language model. This could be acceptable in a setting like BabyAI where the vocabulary is very small and consists of very simple words but we might see an improvement when we move to more complex environments with a larger vocabulary.

\end{document}